%% file: neurips_2026.tex
\documentclass{article}

\PassOptionsToPackage{numbers, compress}{natbib}

 \usepackage[main, final]{neurips_2026}

\usepackage[utf8]{inputenc} 
\usepackage[T1]{fontenc}    
\usepackage{hyperref}       
\usepackage{url}            
\usepackage{booktabs}       
\usepackage{amsfonts}       
\usepackage{nicefrac}       
\usepackage{microtype}      
\usepackage{xcolor}         
\usepackage{multirow}
\usepackage{enumitem}

\usepackage[textsize=tiny]{todonotes}
\usepackage{subcaption}
\usepackage{graphicx}
\usepackage[most]{tcolorbox}

\usepackage{hyperref}

\usepackage{algorithm}
\usepackage{algorithmic}

\usepackage{amsmath}
\usepackage{amssymb}
\usepackage{mathtools}
\usepackage{amsthm}
\usepackage{xcolor}
\usepackage{tcolorbox}

\usepackage{wrapfig}

\usepackage[capitalize,noabbrev]{cleveref}

\theoremstyle{plain}
\newtheorem{theorem}{Theorem}[section]
\newtheorem{fact}{Fact}[section]

\newtheorem{lemma}[theorem]{Lemma}

\theoremstyle{definition}
\newtheorem{definition}[theorem]{Definition}
\newtheorem{assumption}[theorem]{Assumption}
\newtheorem{remark}[theorem]{Remark}

\definecolor{plotblue}{RGB}{70,105,215}
\definecolor{plotlightblue}{RGB}{239,246,255}

\newtcolorbox{question}[1][]{
  enhanced,
  breakable,
  colback=plotlightblue,
  colframe=plotblue,
  boxrule=0pt,
  leftrule=3pt,
  arc=2pt,
  left=2pt,
  right=2pt,
  top=2pt,
  bottom=2pt,
  fonttitle=\bfseries,
  halign=center,
  #1
}

\title{Differentially Private Sparse Reward Estimation with Preference Feedback}

\author{%
  Meng Ding \\
  Department of Computer Science\\
  University of Massachusetts Boston\\
  Boston, MA, USA\\
  \And
  Mingxi Lei \\
  Department of Computer Science and Engineering\\
  University at Buffalo, SUNY\\
  Buffalo, NY, USA\\
  \And
  Jie Zhang \\
  ETH Zürich\\
  Zürich, Switzerland\\
  \And
  Jinyan Liu \\
  Beijing Institute of Technology\\
  Beijing, China\\
  \And
  Di Wang\thanks{Corresponding author.} \\
  King Abdullah University of Science and Technology (KAUST)\\
  Thuwal, Saudi Arabia
}
\begin{document}

\maketitle

\begin{abstract}
Reinforcement Learning with Human Feedback (RLHF) has become a central paradigm for aligning Large Language Models with human values. However, the reliance on sensitive preference data necessitates rigorous differential privacy (DP) guarantees to prevent user information leakage. Existing private alignment methods typically assume dense reward parameterizations, resulting in estimation error bounds that scale polynomially with the ambient {model size $d$}. Such dependence is prohibitive in modern {overparameterized models} where $d$ is vast. 
In this work, we study \emph{Private Sparse RLHF} problem to address this limitation, assuming that human preferences (reward model) are governed by a subset of $s^* \ll d$ relevant features. 
We provide the first comprehensive theoretical framework for differentially private sparse reward estimation from pairwise preference feedback under the Bradley-Terry-Luce model, and introduce efficient algorithms for both Sample-DP and Label-DP settings. 
Theoretically, under the squared $\ell_2$ parameter estimation error, we establish upper bounds of $\widetilde{O}(s^*/n + {s^*}^2/n^2\varepsilon^2)$ for $(\varepsilon,\delta)$-Sample-DP and $\widetilde{O}(s^*/n + s^*/n\varepsilon^2)$ for $\varepsilon$-Label-DP based on randomized response, where $n$ is the training data size. For the lower bounds, we show that the Sample-DP rate is nearly optimal, and that the guarantee for our Label-DP method cannot be further improved within the randomized-response framework. We also provide a general lower bound for the problem with Label-DP.
Empirically, our framework demonstrates superior efficacy over dense baselines across sentiment generation and safety alignment benchmarks, offering a significantly more favorable privacy-utility trade-off.


\end{abstract}

\vspace{-0.1in}\section{Introduction}\vspace{-0.1in}





Large Language Models (LLMs) have demonstrated remarkable capabilities in generalization and reasoning, yet their pre-training on vast, unfiltered corpora often results in behaviors that are misaligned with harmful or biased content. To address this, Reinforcement Learning from Human Feedback (RLHF) has emerged as the standard paradigm for steering LLMs to align pretrained models with human values and objectives \citep{christiano2017deep,stiennon2020learning,ouyang2022training,bai2022training}. In the typical RLHF pipeline, a reward model is first trained on a dataset of pairwise human preferences (e.g., response $y_1$ is preferred over $y_2$) to approximate the underlying human utility function. This learned reward model then guides the optimization of the language model policy via algorithms such as Direct Preference Optimization (DPO) \citep{rafailov2023direct}.


Despite its success, the reliance on human feedback introduces significant privacy vulnerabilities. The preference datasets used to train reward models often contain sensitive information, ranging from personal opinions to implicit biases, which can be memorized by the model. Consequently, deploying these models risks exposing user data through privacy attacks \citep{carlini2019secret,carlini2021extracting,duan2023flocks,wu2023privately}. To mitigate these risks, recent works have integrated Differential Privacy (DP)\cite{dwork2014algorithmic}, which requires that the private trained model (or its output distribution) does not significantly change when any single entry in the training data is altered, into the alignment process \citep{wu2023privately,chowdhury2024differentially, zhou2025unified,zhang2025towards} (refer to \cref{sec:related_work} for further discussion).

\begin{table*}[t]
\centering
\caption{Estimation error rates for private RLHF in non-sparse and sparse settings with error metric $\|\cdot\|_2^2$. 
Here $d$ is the ambient dimension, $s^*$ is the sparsity level, $n$ is the sample size, and $(\varepsilon,\delta)$ are the privacy parameters. $\dagger$ indicates that the lower bound is for the randomized-response mechanism, and $\ddagger$ denotes a general central Label-DP lower bound for arbitrary Label-DP mechanisms. {$\S$ denotes a best-case upper bound and $\P$ indicates lower bound only for Label-DP.}
}
\label{tab:error_rates}
\small{
\begin{tabular}{lccccc}
\toprule
& Sample-DP & Label-DP & Upper Bound & \multicolumn{2}{c}{Lower Bound}   \\
\midrule
{Sparse (\textbf{Ours})} 
& {\checkmark} 
& 
& $\widetilde{O}\!\left(
        \frac{s^*}{ n}
        +
        \frac{(s^*)^2}{ n^2\varepsilon^2}
    \right)$ 
& $\widetilde{\Omega}\!\left(
        \frac{s^*}{ n}
        +
        \frac{(s^*)^2}{ n^2\varepsilon^2}
    \right)$ \\
\addlinespace

{Sparse (\textbf{Ours})} 
& 
& {\checkmark} 
& $\widetilde{O}\!\left(
        \frac{ s^*}{ n}
        +
        \frac{ s^*}{ n\varepsilon^2}
    \right)$ 
& $\widetilde{\Omega}\!\left(
        \frac{s^*}{ n}
        +
        \frac{s^*}{ n\varepsilon^2}
    \right) \dagger$
& $\widetilde{\Omega}\!\left(
    \frac{s^*}{ n}
    +
    \frac{s^*}{ n^2\varepsilon^2}
\right) \ddagger$\\
\addlinespace

{Non-Sparse \cite{chowdhury2024differentially}}  
& {\checkmark}  
& {\checkmark}  
& $\widetilde{O}\!\left(
        \frac{d^2}{n}
        +
        \frac{d^3}{n^2\varepsilon^2}
    \right) \S $ 
& $\widetilde{\Omega}\!\left(
        \frac{d}{n}
        +
        \frac{d}{n^2\varepsilon^2}
    \right) \P$ \\
\bottomrule
\end{tabular} 
\vspace{-0.1in}
}
\end{table*}

However, a fundamental limitation of existing private alignment methods is their implicit treatment of the reward model as a dense parameter vector. In practice, collecting high-quality human preference data for aligning large language models is both costly and time-consuming \citep{chen2022human, mahan2024generative, yao2025leveraging}. For instance, aligning models with general human values often requires a substantial number of annotations from experienced annotators to reliably distinguish between “helpful” and “harmless” responses. This data scarcity becomes even more pronounced in specialized domains such as medical or legal reasoning \citep{singhal2023large, cui2023chatlaw,dong2025understanding}, where preference labels must be provided by highly qualified experts. These challenges impose a strict budget on the available sample size $n$, while the semantic richness and complexity of such domains naturally induce a feature space  of extremely high dimension $d$. Standard differentially private mechanisms require noise injection proportional to the ambient dimension (i.e., $\propto d/n$) to hide individual contributions \citep{dwork2014algorithmic}. When $d \gg n$, this noise overwhelms the utility signal, rendering the high estimation error and alignment ineffective.

Moreover, recent empirical studies \citep{aghajanyan2021intrinsic,christopoulou2024sparsepo} suggest that the underlying reward functions and feature representations in LLM alignment exhibit significant intrinsic sparsity. Specifically, preference signals are often governed by a small subset of salient features or tokens, rather than the full high-dimensional embedding. In parallel, foundational theoretical work on differentially private sparse learning has shown that, for problems such as linear regression, one can circumvent polynomial dependence on the ambient dimension $d$ \citep{talwar2015nearly, wang2019sparse, cai2021cost,ding2024revisiting,zhu2024truthful,tian2025differentially,ding2025nearly}. By leveraging sparsity, the estimation error can be reduced to scale logarithmically with the sparsity (i.e., $\propto s^* / n$), thereby enabling accurate private estimation even in high-dimensional settings. This naturally raises the following question:

\begin{question}
    Can we design effective mechanisms for \emph{private sparse RLHF} with rigorous theoretical guarantees {that are not polynomial in the dimension $d$}?
\end{question}

We answer this question affirmatively and summarize our contributions as follows:
\begin{itemize}[leftmargin=*]
    \item \textbf{First Analysis of Private Sparse RLHF.} To the best of our knowledge, this is the first work to investigate sparsity in the context of private alignment, {i.e., DP Sparse RLHF}. We formulate the problem under the high-dimensional Bradley-Terry model and introduce two specialized algorithms: Sample-DP-IHT, which integrates iterative hard thresholding with a private peeling mechanism to protect the entire interaction tuple, and Label-DP RR+IHT, which combines randomized response with debiased gradient estimation to protect preference labels.

    \item \textbf{Minimax Optimality.} As summarized in Table~\ref{tab:error_rates}, we establish upper bounds under the squared $\ell_2$ parameter estimation error. Specifically, we achieve rates of $\widetilde{O}(s^*/n + (s^*)^2/(n^2\varepsilon^2))$ for $(\varepsilon,\delta)$-Sample-DP and $\widetilde{O}(s^*/n + s^*/(n\varepsilon^2))$ for $\varepsilon$-Label-DP, where $n$ is the training data size. For the lower bounds, we show that the Sample-DP rate is nearly optimal, and that our theoretical analysis for the Label-DP method cannot be further improved. We also provide a general lower bound $\widetilde{\Omega}\!\left(
    \frac{s^*}{ n}
    +
    \frac{s^*}{ n^2\varepsilon^2}
\right) $ for the problem with  Label-DP.
     
    \item \textbf{Experimental Evaluation.} We empirically validate our framework on Sentiment Generation and Safety Alignment benchmarks. Our results demonstrate that exploiting sparsity consistently yields superior privacy-utility trade-offs compared to dense baselines (e.g., DP-SGD and Randomized Response), effectively mitigating the noise overhead in high-privacy regimes and achieving significantly higher reward scores and alignment rates.
\end{itemize}

\vspace{-0.1in}\section{Related Work}\label{sec:related_work}\vspace{-0.1in}

\noindent\textbf{Differentially Private Reward Learning and Alignment} Recent work has increasingly focused on integrating differential privacy into the alignment of LLMs. \citet{chowdhury2024differentially} investigates differentially private reward estimation using preference feedback, while subsequent studies extend these guarantees to Direct Preference Optimization and unified offline alignment frameworks \citet{chowdhury2024provably, zhou2025unified, chen2025improved}. Beyond reward estimation, privacy-aware regularization has received growing attention. \citet{wu2025offline} analyzes offline and online KL-regularized RLHF under differential privacy, while \citet{zhang2025kl} shows that KL regularization itself can inherently induce differential privacy guarantees in bandit and RLHF settings. The scope of private alignment has further expanded to more complex interaction models, such as reinforcement learning with self-play \cite{qiao2024differentially} and user-level differential privacy in RLHF pipelines \cite{zhang2025towards}. Despite these advancements, prior works largely focused on dense reward structures or general function approximation, whose the estimation errors always depend on polynomial of the dimension, leaving the sparse reward learning under privacy constraints unexplored.

\noindent\textbf{Differentially Private Sparse Learning}
Differentially private sparse learning builds on a broader line of work on private empirical risk minimization (ERM) and convex optimization \citep{chaudhuri2011differentially,kifer2012private,huai2020pairwise,wang2019noninteractive,wang2020empirical,DBLP:journals/corr/abs-2505-15647}. Early studies \citep{thakurta2013differentially} investigated the stability of the Lasso estimator, clarifying the conditions required for robust feature selection under privacy constraints. Building on these foundations, \citet{talwar2015nearly} gives near-optimal utility guarantees for differentially private Lasso, establishing sharp statistical rates (up to logarithmic factors) in sparse linear regression under central DP.
More recent work has expanded both the algorithmic toolbox and the problem settings. \citet{wang2020knowledge} propose a knowledge-transfer framework that can improve utility for private sparse regressions. \cite{tao2022private,hu2022high} addressed private stochastic convex optimization under heavy-tailed gradients by robust estimation and median techniques. In parallel, sparse learning under local differential privacy, \citet{wang2019sparse} develops LDP iterative hard thresholding procedures, and subsequent analysis \citet{zhu2023improved} sharpen lower/upper bounds and provide the first non-interactive LDP algorithms for the problem. Beyond regression, privacy-preserving sparse spectral estimation has also been studied. \citet{hu2023privacy,hu2023nearly,wang2020principal} examined the sparse generalized eigenvalue problem underlying sparse principal component analysis, canonical correlation analysis, and Fisher’s discriminant analysis.
Despite these advances, existing private sparse learning results can not straightforwardly extend to the {preference-based learning} setting central to RLHF, along with the non-linear Bradley-Terry-Luce model and paired data dependency. 

\vspace{-0.1in}\section{Preliminaries}\label{sec:preliminaries}
In this section, we present the formal problem setup for private alignment in RLHF from pairwise comparisons, along with the necessary definitions and assumptions. \vspace{-0.1in}

Let $D=\{(k_i,a_i^0,a_i^1,y_i)\}_{i=1}^n$ be a dataset consisting of $n$ labeled comparisons drawn i.i.d. from $\mathcal{D}$, where $k_i\in\mathcal{K}$ denotes a context (e.g., a prompt or state), $a_i^0,a_i^1\in\mathcal A$ consists of two candidate actions/responses, and the label $y_i\in\{0,1\}$ indicates which candidate is preferred (i.e., $y_i=l$ means that $a_i^l$ is chosen as the preferred response). Let $r_\theta:\mathcal{K}\times\mathcal A\to\mathbb R$ be a parametric reward model indexed by $\theta\in\Theta$, and let $\theta^\star$ denote the unknown ground-truth parameter. We assume the preferences are generated according to the softmax comparison model: for each $i\in[n]$ and $l\in\{0,1\}$,
$$
\begin{aligned}
    \mathbb{P}_{\theta^\star}\!\left[y_i=l \,\middle|\, k_i,a_i^0,a_i^1\right]= \frac{\exp \left(r_{\theta^\star}\left(k_i, a_i^l\right)\right)}
{\exp \left(r_{\theta^\star}\left(k_i, a_i^0\right)\right)+\exp \left(r_{\theta^\star}\left(k_i, a_i^1\right)\right)},
\end{aligned}
$$
where $r_{\theta^*}(\cdot, \cdot)$ denotes the reward model parameterized by $\theta^*$. This model is known as the Bradley-Terry-Luce (BTL) model \cite{bradley1952rank,luce1959individual}. In this work, we focus on the BTL model with a linear reward parameterization. Specifically, let $\phi:\mathcal{K}\times\mathcal A\to\mathbb R^d$ be a known feature map and assume $r_\theta(k,a)=\langle \phi(k,a),\theta\rangle, \theta\in\mathbb R^d.$
Define the comparison (difference) feature $x_i:=\phi(k_i,a_i^1)-\phi(k_i,a_i^0)\in\mathbb R^d.$
Under this model, the conditional distribution reduces to logistic regression:
$$
\begin{aligned}
    \mathbb{P}_{\theta^\star}\!\left[y_i=1 \,\middle|\, k_i,a_i^0,a_i^1\right]=\sigma\!\left(\langle x_i,\theta^\star\rangle\right), 
    \mathbb{P}_{\theta^\star}\!\left[y_i=0 \,\middle|\, k_i,a_i^0,a_i^1\right]=\sigma\!\left(-\langle x_i,\theta^\star\rangle\right),
\end{aligned}
$$
where $\sigma(z)=\frac{1}{1+e^{-z}}$ refers to the sigmoid function. 

Given $D$, our objective is to estimate $\theta^\star$ (and hence the reward function $r_{\theta^\star}$) by maximizing the likelihood of the observed comparisons with privacy constraints. Formally, we seek a randomized estimator $\mathcal{M}: D \to \mathbb{R}^d$ that minimizes the empirical risk $\mathcal{L}_n(\theta)$ subject to the privacy constraint, where $\mathcal{L}_n(\theta)$ is defined as follows:
$
\mathcal{L}_n(\theta)=\frac{1}{n}\sum_{i=1}^n\Big(\log(1+\exp(\langle x_i,\theta\rangle)) - y_i\langle x_i,\theta\rangle\Big).
$
To account for the high-dimensional nature of the feature space ($d \gg n$), we adopt the following assumptions, which are standard in the literature \cite{shah2016estimation,chowdhury2024differentially,zhou2025unified,yao2025leveraging}. 

\begin{assumption}[Boundedness]\label{asm:boundness}
    (a) $\theta^*$ lies in the set $\Theta_{B, s^*}=\left\{\theta \in \mathbb{R}^d \mid \|\theta\|_2 \leq B, \|\theta\|_0 \leq s^* \right\}$. (b) Features are bounded, i.e., $\|x\|_{\infty} \leq 1, \forall x=\phi(k,a^1)-\phi(k,a^0)$. (c) Let $\Sigma:=\mathbb{E}[x x^{\top}]$, and suppose there exist constants $0<\mu \leq \gamma<\infty$ such that $\mu \leq \lambda_{\min }(\Sigma) \leq \lambda_{\max }(\Sigma) \leq \gamma$.
\end{assumption}

Our boundedness assumptions differ from the $\ell_2$ feature bound used in \cite{chowdhury2024differentially}. {Note that in the high dimensional setting, $\ell_\infty$-boundness assumption is more reasonable than $\ell_2$-norm. Such an assumption has been widely used in the existing literature on high-dimensional statistics~\cite{vershynin2018high}}. Under an $\ell_2$ bound
$\|x\|_2\le L$, one has $\operatorname{tr}(\Sigma)=\mathbb E\|x\|_2^2\le L^2,$ and hence $\lambda_{\min}(\Sigma)\le \frac{L^2}{d}.$
Thus, when $L=O(1)$, a dense design cannot have all covariance eigenvalues bounded below by a constant; the curvature parameter $\mu$ may scale as $1/d$, leading to dimension dependence through the condition number. In our work, we assume the coordinate-wise bound $\|x\|_\infty\le 1$, which allows $\|x\|_2=O(\sqrt d)$ and permits covariance eigenvalues $\mu,\gamma=\Theta(1)$. We will provide more discussions in the later parts.

Now, we recall the definition of differential privacy \cite{dwork2014algorithmic}. Informally, it ensures that the privately trained model (or its output distribution) does not significantly change when any single entry in the training data is altered.
Depending on the scenario, there are two levels of privacy: label-DP \cite{chaudhuri2011sample,ghazi2021deep} and sample-DP \cite{chaudhuri2011differentially,dwork2014algorithmic,abadi2016deep}. We provide their formal definitions in the following.
\begin{definition}[Differential Privacy \cite{dwork2014algorithmic}]\label{def:central_dp}
Let $\varepsilon \ge 0$ and $\delta \in (0,1]$. A randomized algorithm $\mathcal{M}$ is said to satisfy \emph{$(\varepsilon,\delta)$-differential privacy} with respect to an adjacency relation $\sim$, if for any pair of adjacent datasets $D \sim D'$, and for any measurable set $S$ in the output space of $\mathcal{M}$, it holds that
$$
\mathbb{P}[\mathcal{M}(D) \in S] \leq e^{\varepsilon} \cdot \mathbb{P}\left[\mathcal{M}\left(D^{\prime}\right) \in S\right]+\delta.
$$
When $\delta=0$, we say that $\mathcal{M}$ is {$\varepsilon$-differentially private}. Throughout the paper, we consider two specific types of adjacency: 
\textbf{1) Sample Adjacency (Sample-DP).} $D$ and $D'$ differ in exactly one sample record (i.e., replacement or addition/removal of a pair $(x_i, y_i)$). \textbf{2) Label Adjacency (Label-DP).} $D$ and $D'$ differ only in the label of a single sample (i.e., fixed features $x_i$, different $y_i$).
\end{definition}
Definition \ref{def:central_dp} guarantees that the trained reward model remains statistically stable regardless of variations in a single user's data. Particularly, Label-DP ensures that an adversary cannot determine the specific preference $y_i$ for a fixed pair of responses. In contrast, Sample-DP provides a strictly stronger guarantee by protecting the entire interaction tuple $(k_i, a_i^0, a_i^1, y_i)$, concealing not just the preference but also the context of the query itself. Notably, prior works on private RLHF have mainly focused on the Label-DP setting \citep{chowdhury2024differentially, zhou2025unified}, leaving the theoretical analysis of Sample-DP largely unexplored even under non-sparse settings.

\vspace{-0.1in}\section{Differentially Private Sparse RLHF}\vspace{-0.1in}

In this section, we develop sparse reward estimators with differential privacy guarantees under two privacy regimes: Sample-DP and Label-DP. 
For Sample-DP, under our BTL model and Assumption~\ref{asm:boundness}, we establish nearly matching upper and lower bounds, demonstrating minimax optimality up to logarithmic factors. 
For Label-DP, we focus on the randomized-response label privatization paradigm and show that our RR-based sparse estimator cannot be further improved within this framework. We further provide a general lower bound for arbitrary central Label-DP mechanisms.

\vspace{-0.1in}\subsection{Sample-DP} \label{sec:sample_dp}
To achieve sample-level differential privacy, a natural approach is to incorporate the Gaussian mechanism, for instance via DP-SGD \cite{abadi2016deep}. However, in high-dimensional sparse settings, directly applying DP-SGD may lead to suboptimal utility, as the injected noise scales with the ambient dimension and may not provide an efficient estimation of the sparse parameter.
Therefore, we introduce a differentially private variant of the iterative hard thresholding (IHT) algorithm. Algorithm~\ref{alg:sample_dp_iht} begins with a sample-splitting strategy, wherein disjoint subsets of the data are used for successive gradient updates across iterations. We then replace the deterministic hard thresholding step with a differentially private \emph{Peeling Mechanism} \cite{cai2021cost}. At each iteration, the peeling mechanism privately selects the indices corresponding to the largest $s$ entries in magnitude and retains only these coordinates, setting all remaining entries to zero. Refer to \cref{alg:sample_dp_iht} and \cref{alg:peeling} for details.

\begin{algorithm}[tb]
   \caption{Sample-DP-IHT}
   \label{alg:sample_dp_iht}
\begin{algorithmic}[1]
   \STATE {\bfseries Input:} Dataset ${D}=\{(y_i,z_i)\}_{i\in[n]}$, step size $\eta$, sparsity level $s$, privacy parameters $(\varepsilon,\delta)$, total steps $T$.
   \STATE {\bfseries Initialize:} $\theta^{0} \leftarrow 0$.
   
   \STATE \COMMENT{\textit{Step 1: Sample Splitting}}
   \STATE Split ${D}$ into $T$ disjoint parts $\{{D}_t\}_{t=1}^{T}$, each containing $m=n/T$ samples.

   \STATE \COMMENT{\textit{Step 2: Iterative Optimization}}
   \FOR{$t=0$ {\bfseries to} $T-1$}
       \STATE Compute the gradient update:
       $\quad \theta^{t+\frac12} \leftarrow \theta^{t} - \frac{\eta}{m}\sum_{i\in {D}_{t}} \nabla \ell (z_i,y_i;\theta^t)$
       
       \STATE Apply Private Hard Thresholding (Algorithm~\ref{alg:peeling}):
       $$\quad \theta^{t+1} \leftarrow \Pi_{B}\left(\mathrm{Peeling}\left(\theta^{t+\frac12}, s, \varepsilon, \delta, \frac{\eta}{m}\right)\right),$$ where $\Pi_B$ is the projection onto the $B$-radius ball
   \ENDFOR
   
   \STATE {\bfseries Output:} $\theta^{T}$
\end{algorithmic}
\end{algorithm}



\begin{theorem}\label{thm:sample_dp_upper_bound}
    For any $0<\varepsilon, \delta<1$, Algorithm~\ref{alg:sample_dp_iht} satisfies $(\varepsilon, \delta)$-Sample-DP. Under Assumption~\ref{asm:boundness}, suppose that the initialization $\theta^0$ satisfies $\left\|\theta^0- \theta^*\right\| \leq O\left(\frac{B \mu}{\gamma}\right)$. Assume further that the sample size $n$ is sufficiently large such that $n^2 \geq O\left(\frac{\gamma}{\mu m(B)} \frac{s^2 T^2 \log ^2 \frac{s}{\zeta} \log \frac{1}{\delta}}{ B\varepsilon^2}\right)$. If we choose $T=\widetilde{O}\left(\frac{\gamma}{\mu} \log n\right)$, $s \asymp s^*$ and $\eta=\frac{8}{3 \gamma}$, then with high probability, it holds that $\|\theta^T-\theta^*\|_2^2
    \leq
    \widetilde O\left(
        \frac{\gamma^2 s^*}{\mu^3 n}
        +
        \frac{\gamma (s^*)^2}{\mu^4 n^2\varepsilon^2}
    \right).$
\end{theorem}
    

\begin{remark}
{Theorem~\ref{thm:sample_dp_upper_bound} shows that our Sample-DP-IHT estimator achieves an error bound that is polynomial in the sparsity $s^*$, instead of ambient dimension $d$. This is the main distinction from prior dense private reward estimation results \cite{chowdhury2024differentially} (we need to  clarify that although their main objective-perturbation result is stated for Label-DP, the same objective perturbation mechanism can naturally extend to Sample-DP. Under their most favorable parameter choices, this yields the best-case upper bound $\widetilde O( \frac{d^2}{n} + \frac{d^3}{n^2\varepsilon^2}).$ ) 
Note that this comparison should not be interpreted by simply setting $s^*=d$. In the dense case, our bound is not directly comparable to that of \cite{chowdhury2024differentially}, because the two analyses use different feature-normalization assumptions. Specifically, \cite{chowdhury2024differentially} assumes an $\ell_2$ feature bound $\|x\|$, whereas our Assumption~\ref{asm:boundness} imposes the coordinate-wise bound $\|x\|_\infty$. These differences lead to different dimension dependence through the curvature and coverage parameters.
}
\end{remark}

In the following, we formally define the minimax risk for private estimation and introduce the Score Attack framework \cite{cai2023score} used to derive our lower bounds.

Recall the dataset $D=\{(x_i, y_i)\}_{i=1}^n$ consisting of $n$ i.i.d. observations, where $x_i \in \mathbb{R}^d$ denotes the difference feature vector (as defined in Section \ref{sec:preliminaries}) drawn from a distribution $\mathcal{D}_x$, and $y_i \in \{0, 1\}$ is the preference label.
Consistent with the BTL assumption, the conditional distribution of $y_i$ given $x_i$ follows the logistic model, which belongs to the canonical exponential family:
\begin{equation}\label{eq:low_data_gene}
    f_{\theta^{\star}}(y_i \mid x_i) = \exp \left( y_i \langle x_i, \theta^{\star} \rangle - \psi(\langle x_i, \theta^{\star} \rangle) \right),
\end{equation}
where $\psi(u) = \log(1+e^u)$ is the log-partition function. 

We aim to estimate the unknown parameter $\theta^*$ using an estimator $\hat{\theta}: (\mathbb{R}^d \times \{0,1\})^n \to \Theta$. The quality of the estimator is measured by the squared $\ell_2$-norm loss $L(\hat{\theta}, \theta^*) = \|\hat{\theta} - \theta^*\|_2^2$. Then, the classical {statistical minimax risk} is defined as follows:
{{
\begin{equation*}
    \mathfrak{M}_n(\Theta_{B, s^*}) := \inf_{\hat{\theta}} \sup_{\theta \in \Theta_{B, s^*}} \mathbb{E}_{D \sim P_\theta} \left[ \|\hat{\theta}(D) - \theta\|_2^2 \right],
\end{equation*}
}}
where $P_\theta$ denotes the joint distribution of the dataset induced by the parameter $\theta$.

In the $(\varepsilon, \delta)$-differential privacy model, the estimator must be obtained via a randomized mechanism $\mathcal{M}$ (as defined in Section \ref{sec:preliminaries}). Let $\mathcal{Q}_{\varepsilon, \delta}$ denote the set of all $(\varepsilon, \delta)$-DP mechanisms. The {$(\varepsilon, \delta)$-private minimax risk} is defined as: $\mathfrak{M}_{n, \varepsilon, \delta}(\Theta_{B, s^*}) := \inf_{\mathcal{M} \in \mathcal{Q}_{\varepsilon, \delta}} \sup_{\theta \in \Theta_{B, s^*}} \mathbb{E}_{\mathcal{M}, D \sim P_\theta} \left[ \|\mathcal{M}(D) - \theta\|_2^2 \right].$

To derive tight lower bounds for the private minimax risk, we utilize the \textit{Score Attack} framework \citep{cai2023score}. Given Eq.~\eqref{eq:low_data_gene}, the {score statistic} for a data point $z_i = (x_i, y_i)$ is defined as $S(z_i; \theta) := (y_i - \sigma(\langle x_i, \theta \rangle)) x_i$. This statistic is centered with variance equal to the Fisher information matrix $\mathcal{I}(\theta)$. The \textit{Score Attack statistic} $\mathcal{A}_{\theta}$ is defined as the correlation between the estimation error and the score function: $\mathcal{A}_{\theta}(z, \mathcal{M}(D)) := \langle \mathcal{M}(D) - \theta, S(z; \theta) \rangle.$

The Score Attack captures the fundamental tension between utility and privacy. Intuitively, a good estimator must react to the signal in the training data, leading to a high total correlation with the score function on $D$. 
However, differential privacy strictly limits this correlation to prevent membership inference, which is quantified by the following lemma:

\begin{algorithm}[tb]
   \caption{Peeling Procedure}
   \label{alg:peeling}
\begin{algorithmic}[1]
   \STATE {\bfseries Input:} Vector $\theta\in\mathbb{R}^d$, sparsity $s$, privacy parameters $(\varepsilon,\delta)$, noise scale $\lambda$.
   \STATE {\bfseries Initialize:} $S \leftarrow \emptyset$.
   
   \STATE \COMMENT{\textit{Phase 1: Iterative Support Selection}}
   \FOR{$i=1$ {\bfseries to} $s$}
       \STATE Generate noise vector $v_i \in \mathbb{R}^d$ where each component is drawn from:
       \STATE $\quad v_{i,j} \sim \mathrm{Lap}\left(\frac{2\lambda\sqrt{3k\log(1/\delta)}}{\varepsilon}\right), \quad \forall j \in [d]$
       
       \STATE Select the index with the maximum noisy magnitude:
       \STATE $\quad j^\star \leftarrow \arg\max_{j\in[d]\setminus S} \left(|\theta_j| + v_{i,j}\right)$,
       $\quad S \leftarrow S \cup \{j^\star\}$
   \ENDFOR
   
   \STATE \COMMENT{\textit{Phase 2: Final Parameter Perturbation}}
   \STATE Generate output noise vector $\widetilde{v} \in \mathbb{R}^d$:
   \STATE $\quad \widetilde{v}_j \sim \mathrm{Lap}\left(\frac{2\lambda\sqrt{3k\log(1/\delta)}}{\varepsilon}\right), \quad \forall j \in [d]$
   
   \STATE {\bfseries Output:} $\theta \!\mid_{S} + \widetilde{v}_{S}$
\end{algorithmic}
\end{algorithm}

\begin{lemma}[Privacy Bound on Score Attack \citep{cai2023score}]\label{lem:privacy_bound}
    If $\mathcal{M}$ satisfies $(\varepsilon, \delta)$-differential privacy with $0 < \varepsilon < 1$, then for any $\theta \in \Theta$, it holds that $\left| \sum_{i=1}^n \mathbb{E}[\mathcal{A}_{\theta}(z_i, \mathcal{M}(D))] \right| \le n\varepsilon \sqrt{\lambda_{\max}(\mathcal{I}(\theta))} \sqrt{\mathbb{E}\|\mathcal{M}(D) - \theta\|_2^2} + 2n\delta s^* \sqrt{\log(1/\delta)}.$
\end{lemma}

By combining the statistical necessity of a high score correlation (for utility) with the privacy upper bound in Lemma~\ref{lem:privacy_bound}, we obtain the following minimax lower bound, where full proofs are detailed in Appendix~\ref{sec:app_proofs}.

\begin{theorem}\label{thm:sample_dp_lower_bound}
    Consider i.i.d. observations $(y_1, {x}_1)$,$\cdots$, $(y_n, {x}_n)$ drawn from Eq.\eqref{eq:low_data_gene} with $\theta^* \in \Theta$ and  Assumption~\ref{asm:boundness} holds. Suppose that $s^*$ satisfies $s^* \lesssim \sqrt{d}$, and $s^* \gg \log ^2\left(d / s^*\right)$. If the estimator $\mathcal{M}$ is $(\varepsilon, \delta)$-sample-differentially private with $0<\varepsilon<1$, $s^* \log \left(d / s^*\right) \lesssim n \varepsilon$, and $\delta \lesssim n^{-(1+c)}$ for some $c>0$, then
    $ \inf _{\mathcal{M} \in \mathcal{Q}_{\varepsilon, \delta}} \sup _{{\theta} \in \Theta} \mathbb{E}\|\mathcal{M}(D)-{\theta}\|_2^2 \geq \widetilde{\Omega}\left(\frac{s^* }{\gamma n}+\frac{\left(s^* \right)^2}{\gamma n^2 \varepsilon^2}\right) .$
\end{theorem}

Theorem~\ref{thm:sample_dp_lower_bound} establishes a matching lower bound to Theorem~\ref{thm:sample_dp_upper_bound} up to logarithmic factors (if we omit other factors), showing the nearly minimax optimality of our proposed estimator.
    

\vspace{-0.1in}\subsection{Label-DP}\label{sec:labeldp}
To achieve Label-DP, the privacy requirement is relaxed such that the feature vectors $x_i$ (derived from contexts and prompts) are considered public, while only the preference labels $y_i$ are sensitive. This assumption enables a clean separation between the privatization and optimization stages. Rather than injecting noise into gradients as in the sample-level DP setting, we adopt an input perturbation strategy consisting of three steps: (1) privatizing labels via Randomized Response, (2) constructing unbiased gradient estimators via debiasing, and (3) optimizing the objective using standard Iterative Hard Thresholding.

\noindent\textbf{Step 1: Randomized Response (RR).}
To protect the privacy of the labels $y_i \in \{0, 1\}$, we apply the classical randomized response mechanism $\mathcal{M}_{\text{RR}}$ \citep{warner1965randomized}. For each sample $i$, the mechanism outputs a perturbed label $\widetilde{y}_i$ by flipping the true label with a probability determined by the privacy budget $\varepsilon$:
\begin{equation*}
    \mathbb{P}(\widetilde{y}_i = y_i) = \frac{e^\varepsilon}{e^\varepsilon + 1}, \quad \text{and} \quad \mathbb{P}(\widetilde{y}_i = 1 - y_i) = \frac{1}{e^\varepsilon + 1}.
\end{equation*}
This mechanism guarantees $\varepsilon$-Label DP for each individual response.

\begin{algorithm}[tb]
   \caption{Label-DP RR + IHT}
   \label{alg:rr-debias-iht}
\begin{algorithmic}[1]
   \STATE {\bfseries Input:} Dataset $D=\{(x_i, y_i)\}_{i=1}^n$, sparsity level $s$, privacy budget $\varepsilon$, step size $\eta$, total iterations $T$.
   \STATE {\bfseries Initialize:} $\theta^0 \leftarrow 0$.
   \STATE {\bfseries Parameters:} $p \leftarrow \frac{e^\varepsilon}{e^\varepsilon + 1}$, $q \leftarrow \frac{1}{e^\varepsilon + 1}$.
   
   \STATE \COMMENT{\textit{Step 1: Label Privatization (Randomized Response)}}
   \FOR{$i=1$ {\bfseries to} $n$}
       \STATE Sample $\tilde{y}_i \in \{0, 1\}$ such that:
       $P(\tilde{y}_i = y_i) = p$ and $P(\tilde{y}_i = 1 - y_i) = q$.
   \ENDFOR
   \STATE Let $\widetilde{D} = \{(x_i, \tilde{y}_i)\}_{i=1}^n$ be the privatized dataset.

   \STATE \COMMENT{\textit{Step 2 \& 3: Debiased Gradient Calculation \& IHT}}
   \FOR{$t=0$ {\bfseries to} $T-1$}
       \STATE Compute the empirical gradient on noisy labels:
       $\quad \hat{g}^t \leftarrow \frac{1}{n} \sum_{i=1}^n \nabla \ell(\theta^t; x_i, \tilde{y}_i)$
       
       \STATE Compute the bias correction terms:
       \STATE $\quad g^{-} \leftarrow \frac{1}{n} \sum_{i=1}^n \nabla \ell(\theta^t; x_i, 0)$, $\quad g^{+} \leftarrow \frac{1}{n} \sum_{i=1}^n \nabla \ell(\theta^t; x_i, 1)$
       
       \STATE Construct the unbiased gradient estimator:
       $\quad \widetilde{g}^t \leftarrow \frac{e^\varepsilon + 1}{e^\varepsilon - 1} \left( \hat{g}^t - \frac{g^- + g^+}{e^\varepsilon + 1} \right)$
       
       \STATE Update and Hard Thresholding:
       $\quad \theta^{t+1} \leftarrow \mathcal{H}_{s} \left( \theta^t - \eta \widetilde{g}^t \right)$
   \ENDFOR
   
   \STATE {\bfseries Output:} $\theta^T$
\end{algorithmic}
\end{algorithm}

\noindent\textbf{Step 2: Debiasing Correction.}
Directly optimizing the loss with respect to the perturbed labels $\widetilde{y}_i$ would lead to biased estimates of the reward parameter $\theta^{\star}$. To correct for this, we construct an unbiased gradient estimator $\widetilde{g}$ using the statistical properties of the RR mechanism. Specifically, at iteration $t$, we define the debiased gradient as:
\begin{equation*}\label{eq:debias_gradient}
    \begin{aligned}
        \nabla \mathcal{L}_n^{\text{debias}}(\theta^t) = \widetilde{g}^t &=  \frac{e^{\varepsilon}+1}{e^{\varepsilon}-1} \cdot \left(\hat{g}^t -\frac{{g}^{-}}{e^{\varepsilon}+1} - \frac{{g}^{+}}{e^{\varepsilon}+1} \right), 
    \end{aligned}
\end{equation*}
where $\hat{g}^t = \frac{1}{n} \sum_{i=1}^n \nabla \ell(\theta^t; x_i, \widetilde{y}_i), {g}^{-}  = \frac{1}{n} \sum_{i=1}^n \nabla \ell(\theta^t; x_i, 0), {g}^{+}  = \frac{1}{n} \sum_{i=1}^n \nabla \ell(\theta^t; x_i, 1).$ Here, $\hat{g}^t$ is the empirical gradient computed using the noisy labels, while $g^-$ and $g^+$ represent the gradients assuming the labels are all 0 or all 1, respectively. It is straightforward to verify that $\mathbb{E}_{\mathcal{M}_{\text{RR}}}[\widetilde{g}^t] = \nabla \mathcal{L}_{n}(\theta^t)$, ensuring that this estimator provides an unbiased direction for optimization.

\noindent\textbf{Step 3: Iterative Hard Thresholding (IHT).}
With the privatized labels $\widetilde{{D}} = \{(x_i, \widetilde{y}_i)\}_{i=1}^n$, the problem reduces to a sparse estimation task with noisy but public targets. Since privacy is enforced entirely during the label perturbation phase (Step 1), the subsequent calculation of the debiased gradient and the parameter updates incur no additional privacy cost due to the post-processing property of differential privacy. 
Consequently, unlike the sample-level DP setting, no private support selection mechanism (e.g., peeling) is required inside the loop. We apply the standard iterative hard thresholding algorithm using the debiased gradient estimator:
\begin{equation*}
    \theta^{t+1} \leftarrow \mathcal{H}_s \left( \theta^t - \eta \nabla \mathcal{L}_n^{\text{debias}}(\theta^t) \right),
\end{equation*}
where $\mathcal{H}_s$ is the hard thresholding operator that preserves the $s$ largest entries in magnitude.

\begin{theorem}\label{thm:label_dp_upper_bound}
    For any $0<\varepsilon, \delta<1$, Algorithm~\ref{alg:rr-debias-iht} satisfies $\varepsilon$-Label-DP. Under Assumption~\ref{asm:boundness}, suppose that the initialization $\theta^0$ satisfies $\left\|\theta^0- \theta^*\right\| \leq O\left(\frac{B\mu}{\gamma}\right)$. Assume further that the sample size $n$ is sufficiently large such that $n^2 \geq O\left(\frac{\gamma}{\mu m(B)} \frac{s \log \frac{1}{\delta}}{ \varepsilon^2}\right)$. If we choose $T=\widetilde{O}\left(\frac{\gamma}{\mu} \log n\right)$, $s \asymp s^*$ and $\eta=\frac{8}{3 \gamma}$, then with probability at least $1-\delta$, it holds that
$\|\theta^T-\theta^*\|_2^2
    \leq \widetilde O\left( \frac{\gamma s^*}{\mu^2 n} + \frac{\gamma s^*}{\mu^2 n\varepsilon^2} \right).$
\end{theorem}

\begin{remark}
We compare Theorem~\ref{thm:label_dp_upper_bound} with prior dense private reward estimation results.
\cite{chowdhury2024differentially} study non-sparse reward estimation; in the central Label-DP setting, their objective-perturbation estimator has best-case squared $\ell_2$ rate $\widetilde O(\frac{d^2}{n}+\frac{d^3}{n^2\varepsilon^2})$.
{They also provide RR-based estimators under local Label-DP with a rate of $\widetilde O(\frac{d^2}{n\varepsilon^2})$.}
In contrast, our RR+IHT estimator exploits sparsity and achieves $\widetilde O(\frac{ s^*}{n} +\frac{ s^*}{ n\varepsilon^2}).$
Thus, our central Label-DP bound replaces the ambient-dimension dependence of dense methods by the sparsity $s^*$, up to logarithmic and condition-number factors.
\end{remark}

In the following, we adopt the same problem setup and private minimax risk definition $\mathfrak{M}_{n, \varepsilon, \delta}(\Theta_{B, s^*})$ introduced above. We first present a lower bound for estimators based on randomized-response mechanisms, showing that the results of our RR-based Label-DP method cannot be further improved within this framework. We then provide a general lower bound.




\begin{figure*}[t]
  \centering
  \begin{subfigure}[b]{0.45\textwidth}
    \centering
    \includegraphics[width=\linewidth]{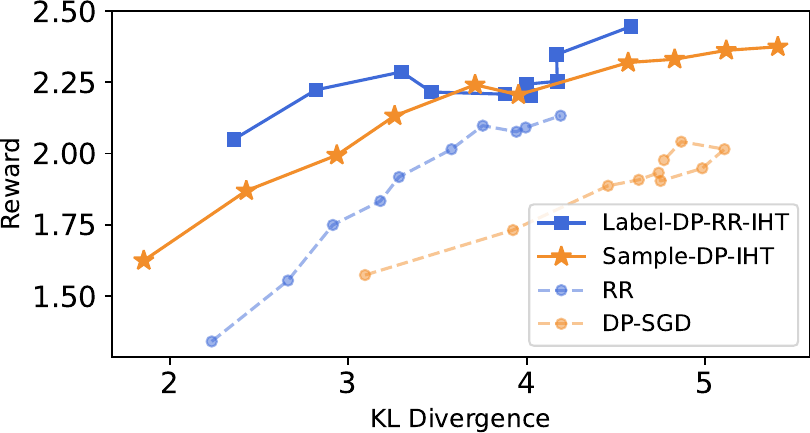}
    \caption{$\varepsilon = 2$}
    \label{fig:kl_reward1}
  \end{subfigure}
  \hfill 
  \begin{subfigure}[b]{0.45\textwidth}
    \centering
    \includegraphics[width=\linewidth]{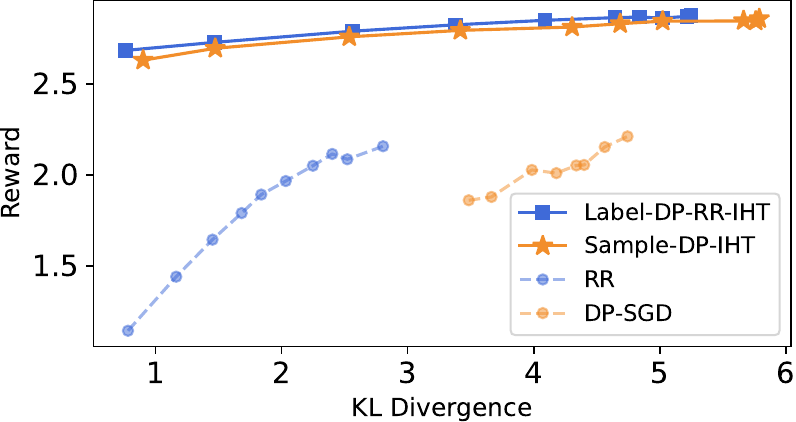}
    \caption{$\varepsilon = 8$}
    \label{fig:kl_reward2}
  \end{subfigure}
  
  \caption{IMDb Sentiment Generation with Llama-7B under different privacy budgets ($\varepsilon$).}
  \label{fig:kl_reward_combined}
  \vspace{-0.1in}
\end{figure*}

\begin{theorem}\label{thm:label_dp_lower_bound}
Consider the BTL model with sparse parameter space $\Theta=\{\theta\in\mathbb R^d:\|\theta\|_2\le B,\ \|\theta\|_0\le s^*\}$ and Assumption \ref{asm:boundness} holds. Then estimator $\mathcal{M}(D)$ based on randomized-response privatized labels $\mathcal{M}_{\text{RR}}$ satisfying central $\varepsilon$-label-DP obeys the minimax lower bound $    \inf_{\mathcal{M}_{RR} \in\mathcal Q_{\varepsilon}}
    \sup_{\theta^*\in\Theta}
    \mathbb{E}\left[\left\|\mathcal{M}_{RR}(D)-\theta^*\right\|_2^2\right] \geq
    \Omega\left(
    \frac{s^*}{\gamma n}
    +
    \frac{s^*}{\gamma n\varepsilon^2}
    \right).$
\end{theorem}

\begin{remark}
{Theorem~\ref{thm:label_dp_lower_bound} establishes the optimality of the randomized-response (RR) label privatization paradigm under the sparse BTL model. Since RR is a local label-DP channel, the result can also be interpreted as a sparse lower bound for estimators based on locally privatized labels. 
It shows that any estimator based on randomized-response privatized labels must incur a privacy-dependent error of order
$\widetilde{\Omega}\left(\frac{s^*}{\gamma n\varepsilon^2}\right)$. This complements the dense local Label-DP results of \cite{chowdhury2024differentially}, whose RR-based estimators and lower bounds are dimension-dependent. 
Our result gives the sparse counterpart, showing that the unavoidable RR privacy cost scales with $s^*$ rather than $d$. 
Therefore, Theorem~\ref{thm:label_dp_upper_bound} is optimal up to logarithmic and condition-number factors within the sparse RR-based framework.}
\end{remark}

\begin{theorem}\label{thm:general_label_dp_lower_bound}
Consider the BTL model with sparse parameter space $\Theta=\{\theta\in\mathbb R^d:\|\theta\|_2\le B,\ \|\theta\|_0\le s^*\}$ and Assumption \ref{asm:boundness} holds. Then any estimator $\mathcal{M}(D)$ satisfying central $\varepsilon$-label-DP obeys the minimax lower bound $\inf_{\mathcal{M}\in\mathcal Q_{\varepsilon}}
    \sup_{\theta^*\in\Theta}
    \mathbb{E}\left[\left\|\mathcal{M}(D)-\theta^*\right\|_2^2\right] \geq
    \Omega\left(
    \frac{s^*}{\gamma n}
    +
    \frac{s^*}{\gamma n^2\varepsilon^2}
    \right).$
\end{theorem}

    Theorem~\ref{thm:general_label_dp_lower_bound} provides a general lower bound for sparse reward estimation under central Label-DP. It shows that, for any central Label-DP mechanism, the squared $\ell_2$ estimation error must incur at least $\Omega(
    \frac{s^*}{\gamma n} + \frac{s^*}{\gamma n^2\varepsilon^2}
    )$.
    However, there remains a gap between this general lower bound and the current upper bound for our Label-DP estimator. Designing Label-DP mechanisms with better privacy dependence beyond randomized response remains an interesting open question.

\vspace{-0.1in}\section{Experiment}\vspace{-0.1in}
In this section, we empirically evaluate our proposed sparse private RLHF algorithms, {Sample-DP-IHT} (Algorithm~\ref{alg:sample_dp_iht}) and {Label-DP RR+IHT} (Algorithm~\ref{alg:rr-debias-iht}). We focus on two distinct alignment tasks: Sentiment  Generation and Safety Alignment. Our primary objective is to demonstrate that explicitly exploiting the sparsity of the reward signal yields a superior privacy-utility trade-off.

\subsection{Experimental Setup}

\noindent\textbf{Tasks and Datasets.} We conduct our evaluation on two standard benchmarks representing common RLHF applications: \textbf{(1) Sentiment Generation (IMDb).} We use the IMDb movie review dataset \citep{maas2011learning}. The goal is to generate text conditioned on a prompt such that the output maximizes positive sentiment. Following established RLHF practices, we treat sentiment classification scores as a proxy for human preference rewards. \textbf{(2) Safety Alignment (Harmful):} We utilize the {Harmful Dataset} provided by the AI2 \footnote{\url{https://huggingface.co/datasets/allenai/real-toxicity-prompts}}. This dataset comprises adversarial prompts designed to elicit harmful or toxic behaviors. The objective is to align the model to safely refuse these requests. We train a safety reward model to penalize harmful compliance and reward safe refusals, evaluating the model's ability to maintain safety under differential privacy constraints.

    

\begin{figure}[t]
\centering
\begin{minipage}[t]{0.48\linewidth}
\vspace{0pt}
\centering
\includegraphics[width=0.92\linewidth]{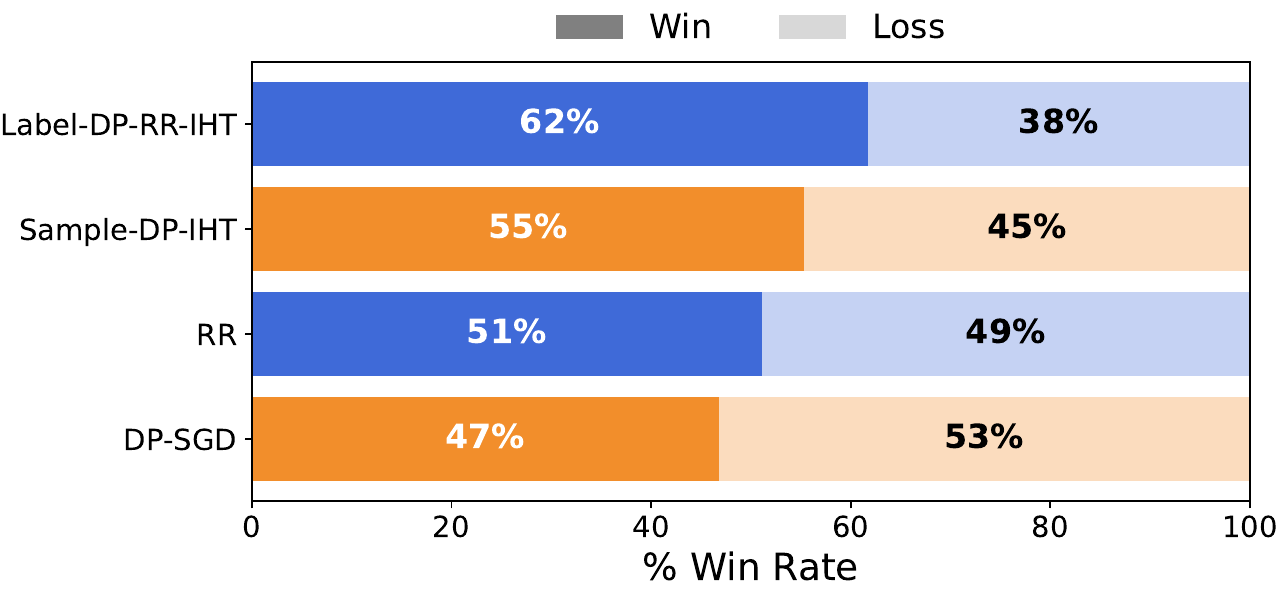}
\vspace{-0.5em}
\captionof{figure}{Safety Alignment with Gemma-3-1B ($\varepsilon=8$).}
\label{fig:safe}
\end{minipage}
\begin{minipage}[t]{0.48\linewidth}
\vspace{0pt}
\centering
\includegraphics[width=0.92\linewidth]{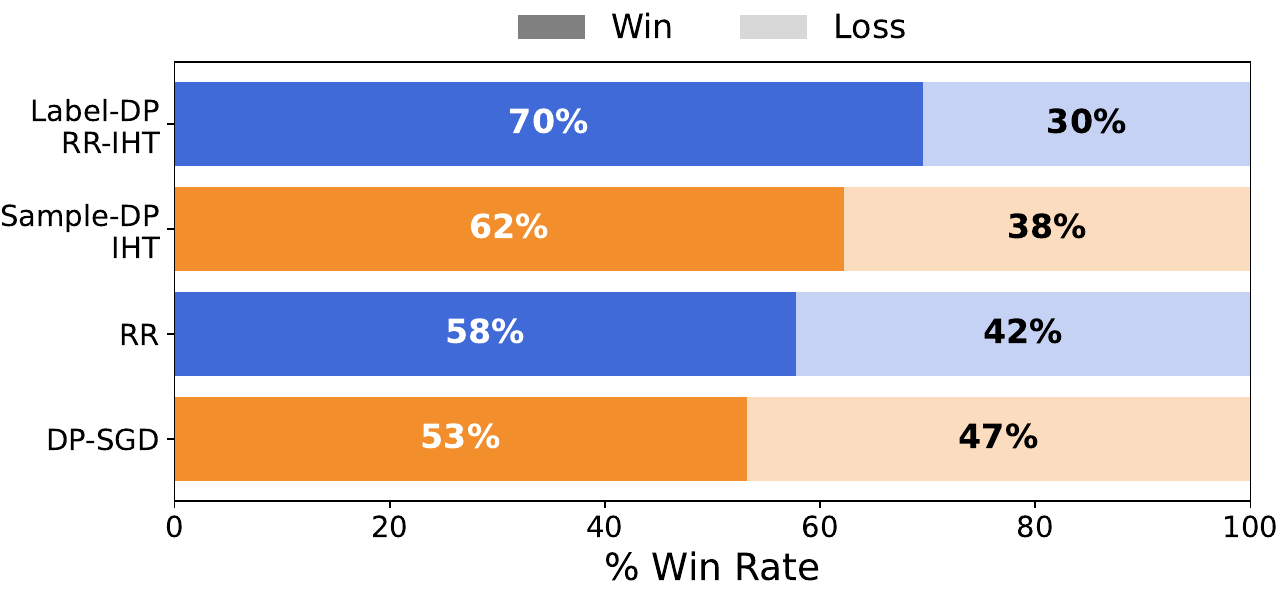}
\vspace{-0.5em}
\captionof{figure}{Safety Alignment with Llama-2-7B ($\varepsilon=8$).}
\label{fig:safe_llama}
\end{minipage}
\vspace{-1em}
\end{figure}

\noindent\textbf{Base Models.} 
We employ two different policy networks to demonstrate generalization across architectures: {LLaMA-7b} \citep{touvron2023llama} for the Sentiment Generation task, and {Gemma-3-1B} \citep{team2025gemma} for the Safety Alignment task. For the reward models, we utilize {RoBERTa} \citep{liu2019roberta} as the backbone architecture. Specifically, we fine-tune separate RoBERTa checkpoints to serve as the sentiment proxy and the safety evaluator, respectively. Unless otherwise noted, all competing methods share identical architectures and initializations to ensure fair comparison.

\noindent\textbf{Baselines.}
We compare our proposed sparse methods against three distinct categories of private alignment baselines for both Sample-DP and Label-DP settings: 
\textbf{(1) DP-SGD:} The standard non-sparse approach \citep{abadi2016deep}. We train the reward model using DP-SGD, clipping per-sample gradients to bound sensitivity and injecting noise proportional to the full dimension $d$. 
\textbf{(2) Randomized Response:} The standard label-differential privacy baseline \citep{chowdhury2024differentially}. We apply the randomized response mechanism to the preference labels without any sparsity-aware thresholding.

\vspace{-0.1in}\subsection{Experimental Results}
\noindent\textbf{Sentiment Generation.}
We visualize the trade-off between reward maximization and KL divergence for the Sentiment Generation task in Figure~\ref{fig:kl_reward1} ($\varepsilon=2$) and Figure~\ref{fig:kl_reward2} ($\varepsilon=8$). Table~\ref{tab:reward} summarizes the peak performance for each method. 
\begin{wraptable}{r}{0.49\textwidth}
\centering
\caption{Performance comparison under different privacy budgets. Higher is better.}
\resizebox{0.46\textwidth}{!}{
\begin{tabular}{lcccc}
\toprule
& DP-IHT & RR-IHT & DP-SGD & RR \\
\midrule
$\varepsilon$ = 2  & 2.37 & 2.44 & 1.98 & 2.12 \\
$\varepsilon$ = 8  & 2.87 & 2.88 & 2.26 & 2.20 \\
\bottomrule
\end{tabular}
}
\label{tab:reward}
\end{wraptable}
The results consistently demonstrate the advantage of sparsity. Across both high-privacy ($\varepsilon=2$) and low-privacy ($\varepsilon=8$) regimes, our sparse algorithms (DP-IHT and RR-IHT) achieve significantly higher rewards for any given KL divergence compared to their dense counterparts (DP-SGD and RR). This indicates that our methods preserve a stronger utility signal, leading to more effective alignment.

\noindent\textbf{Safety Alignment.}
Figures~\ref{fig:safe} and~\ref{fig:safe_llama} present the win rates for the Safety Alignment task under $\varepsilon=8$ using Gemma-3-1B and Llama-2-7B, respectively. 
Across both models, the sparse private methods consistently outperform their dense counterparts. On Gemma-3-1B, Label-DP RR-IHT achieves the highest win rate of \textbf{62\%}, followed by Sample-DP IHT at \textbf{55\%}, while RR and DP-SGD obtain 51\% and 47\%, respectively. On Llama-2-7B, the advantage is even more pronounced: Label-DP RR-IHT reaches \textbf{70\%} and Sample-DP IHT reaches \textbf{62\%}, compared with 58\% for RR and 53\% for DP-SGD. 
This confirms that sparse mechanisms are essential for preventing utility loss in high-dimensional private alignment tasks.

\vspace{-0.1in}\section{Conclusion}\vspace{-0.1in}
In this work, we studied differentially private sparse reward estimation for RLHF under the BTL preference model. By exploiting the intrinsic sparsity of human preferences, we introduced two efficient algorithms: Sample-DP-IHT and Label-DP RR+IHT. Our theory establishes nearly minimax-optimal sparse rates for Sample-DP and shows that the guarantee for RR-based Label-DP cannot be further improved within the randomized-response framework. We also provide a general lower bound for arbitrary central Label-DP mechanisms. Experiments on Sentiment Generation and Safety Alignment demonstrate that sparsity-aware private reward estimation substantially improves the privacy-utility trade-off over dense baselines in the challenging $d\gg n$ regime.

\section*{Acknowledgements}
We thank the anonymous NeurIPS reviewers for their constructive feedback and valuable suggestions, which helped us strengthen and expand this work.


\nocite{langley00}

\bibliography{example_paper}
\bibliographystyle{plainnat}


\appendix
\newpage

\vspace{-0.1in}\section{Addition Experimental Details and Results}

\noindent\textbf{Implementation Details.} We describe the training and evaluation setup for both tasks in the following.
\begin{itemize}
    \item \textbf{Reward Model Training}:
    \sloppy
    For sentiment generation, we initialize the reward model from \emph{cardiffnlp/twitter-roberta-base-sentiment}, which is trained on public Twitter data~\cite{barbieri-etal-2020-tweeteval}. We then adapt it to the IMDb dataset by fine-tuning the classification head. Reward models are trained with differentially private algorithms under $\varepsilon \in \{2, 8\}$ and $\delta = 10^{-5}$. For safety alignment, we follow the same procedure and initialize from \emph{cardiffnlp/twitter-roberta-base-hate}.
    \fussy
    \item \textbf{Proximal Policy Optimization}:
    In the sentiment generation task, we follow the protocol of \citet{rafailov2023direct} and sample prefix prompts with lengths between 2 and 8 tokens. Each approach fine-tunes the base model with PPO using reward signals from its corresponding reward model. In the safety alignment task, we use the toxic prompts in the dataset as prefix prompts and evaluate the models on their generated responses.
    \item
    \textbf{Evaluation}: To ensure comparability across methods, in the sentiment generation, all reward scores are computed with a single fixed evaluator, \emph{siebert/sentiment-roberta-large-english} \citep{hartmann2023}, which we treat as the ground-truth reward model. The reward values are reported on every 20 steps during the training process. For the safety task, we benchmark the algorithms by measuring the win rate against each baseline model. All experiments are conducted on a single NVIDIA A100 GPU.
\end{itemize}

Table~\ref{tab:ppo_params} summarizes the PPO hyperparameters used for the Gemma-3-1B and Llama-2-7B safety alignment experiments. We use the same batch size, mini-batch size, learning rate, optimizer, KL penalty coefficient, and LoRA configuration across the two models. This controlled setup ensures that the observed performance differences mainly come from the reward-estimation methods rather than from different PPO training configurations. In particular, both models are trained with RMSProp, a learning rate of $10^{-5}$, KL penalty coefficient $0.15$, LoRA rank $16$, and LoRA alpha $32$.

\begin{table}[htbp]
\centering
\begin{tabular}{lcc}
\toprule
\textbf{Parameter} & \textbf{Gemma-3-1B} & \textbf{Llama-2-7B} \\
\midrule
Batch Size              & 64      & 64      \\
Mini-batch Size         & 64      & 64      \\
Learning Rate           & 1e-5    & 1e-5    \\
Optimizer               & RMSProp & RMSProp \\
KL Penalty Coefficient  & 0.15    & 0.15    \\
LoRA Rank               & 16      & 16      \\
LoRA Alpha              & 32      & 32      \\
\bottomrule
\end{tabular}
\caption{PPO parameter settings for Gemma-3-1B and Llama-2-7B.}
\label{tab:ppo_params}
\end{table}

\begin{figure}[h]
\centering
\begin{minipage}[t]{0.48\linewidth}
\vspace{0pt}
\centering
\includegraphics[width=0.92\linewidth]{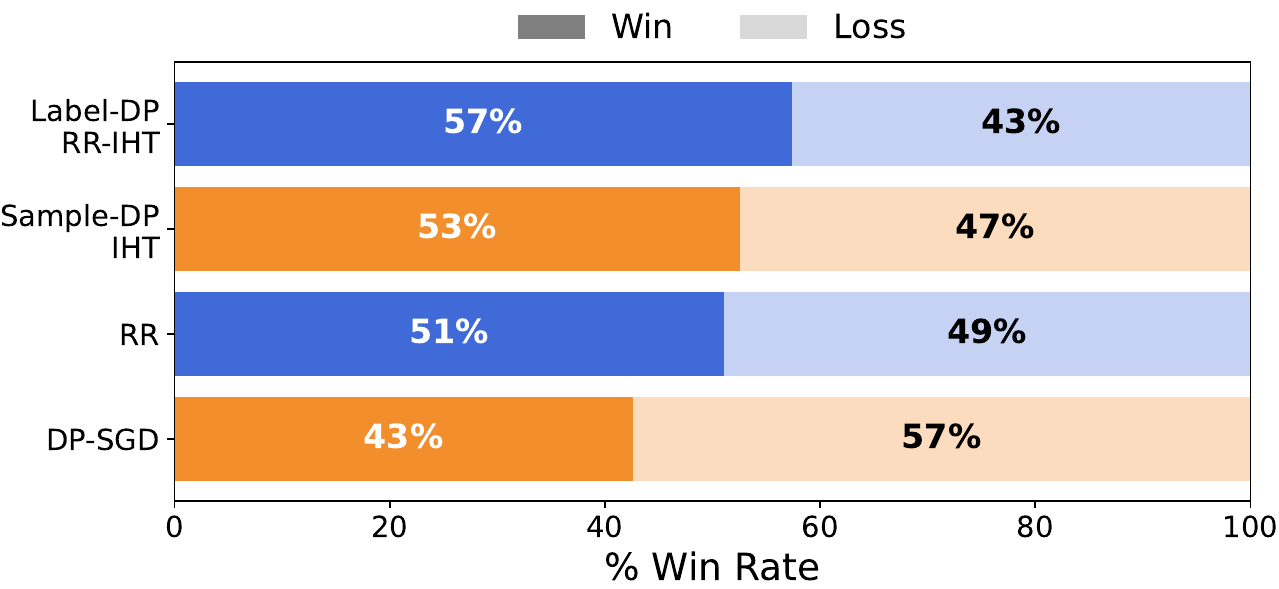}
\vspace{-0.5em}
\captionof{figure}{Safety Alignment with Gemma-3-1B ($\varepsilon=2$).}
\label{fig:safety_gemma_eps2}
\end{minipage}
\begin{minipage}[t]{0.48\linewidth}
\vspace{0pt}
\centering
\includegraphics[width=0.92\linewidth]{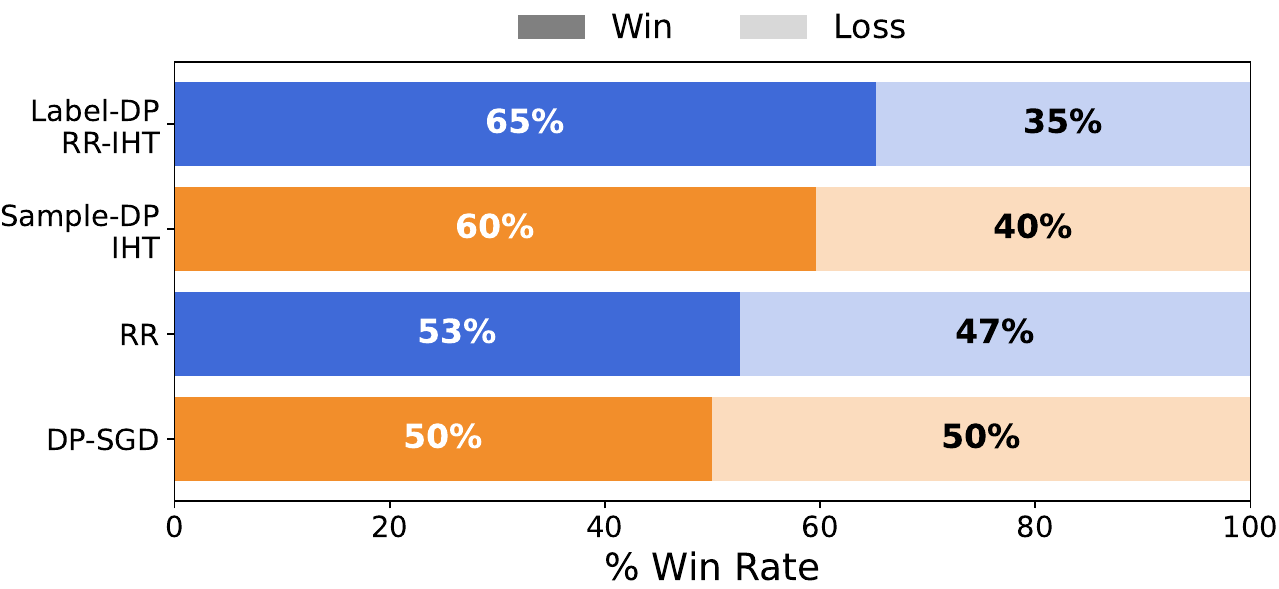}
\vspace{-0.5em}
\captionof{figure}{Safety Alignment with Llama-2-7B ($\varepsilon=2$).}
\label{fig:safety_llama_eps2}
\end{minipage}
\vspace{-1em}
\end{figure}

Figures~\ref{fig:safety_gemma_eps2} and~\ref{fig:safety_llama_eps2} provide additional safety alignment results under a stronger privacy budget $\varepsilon=2$. 
Across both Gemma-3-1B and Llama-2-7B, the sparsity-aware methods consistently outperform their dense counterparts. 
In particular, Label-DP RR+IHT achieves the highest win rates, improving over the dense RR baseline from $51\%$ to $57\%$ on Gemma-3-1B and from $53\%$ to $65\%$ on Llama-2-7B. 
Similarly, Sample-DP-IHT improves over DP-SGD, increasing the win rate from $43\%$ to $53\%$ on Gemma-3-1B and from $50\%$ to $60\%$ on Llama-2-7B. 
These results further confirm that exploiting sparsity helps preserve useful reward signals under strong privacy constraints, leading to better safety alignment performance across different model architectures.

\subsection{Additional Evaluation Protocol Details}
\label{app:rebuttal_protocol}
The sentiment scores are computed by the fixed, independent evaluator
\emph{siebert/sentiment-roberta-large-english}, rather than by the private reward
model used for PPO. The safety-alignment win rates are judged by GPT-4.1.
The evaluators are held fixed across methods and privacy budgets.
We initialize from publicly available pretrained checkpoints and do not train
an SFT stage on sensitive user data. PPO uses the private reward model and public
prompts, so it is post-processing with respect to the private preference-training
data. For the DP-SGD baseline, the clipping norm is $1$, and FastDP calibrates
the Gaussian noise to the target $\varepsilon$ and $\delta$.

\subsection{Non-private Reference and Direct Reward-model Evaluation}
\label{app:rebuttal_nonprivate}
We supplement the private comparisons with a reward model trained without
privacy constraints on the full clean preference dataset and a PPO policy
trained using that reward model. Table~\ref{tab:rebuttal_policy_reference}
extends the sentiment-generation comparison with this non-private reference.
The highest private policy scores are $2.44$ and $2.88$ at $\varepsilon=2$ and
$\varepsilon=8$, respectively, compared with $3.07$ for the non-private
reference, corresponding to absolute score gaps of $0.63$ and $0.19$.

\begin{table}[htbp]
\centering
\caption{Sentiment-generation policy scores with a non-private reference.
Higher is better. The same non-private reference is shown in both rows.}
\label{tab:rebuttal_policy_reference}
\begin{tabular}{lccccc}
\toprule
Privacy budget & Non-private & DP-IHT & RR-IHT & DP-SGD & RR \\
\midrule
$\varepsilon=2$ & 3.07 & 2.37 & 2.44 & 1.98 & 2.12 \\
$\varepsilon=8$ & 3.07 & 2.87 & 2.88 & 2.26 & 2.20 \\
\bottomrule
\end{tabular}
\end{table}

To separate reward estimation from downstream PPO optimization, we also report
direct held-out reward-model performance in
Table~\ref{tab:rebuttal_reward_direct}. DP-IHT and RR-IHT have NLL gaps of
$0.0135$ and $0.0028$ relative to the non-private reference; their preference
accuracies are lower by $0.67$ and $0.04$ percentage points, respectively.

\begin{table}[htbp]
\centering
\caption{Direct reward-model evaluation reported in the rebuttal. Lower mean
loss/NLL and higher held-out preference accuracy are better. }
\label{tab:rebuttal_reward_direct}
\begin{tabular}{lccc}
\toprule
Method & Mean loss/NLL $\downarrow$ & Loss std. & Accuracy $\uparrow$ \\
\midrule
DP-SGD      & 0.5154 & 0.0090 & 0.8302 \\
RR          & 0.4944 & 0.0012 & 0.8568 \\
DP-IHT      & 0.3316 & 0.0008 & 0.8583 \\
RR-IHT      & 0.3209 & 0.0006 & 0.8646 \\
\midrule
Non-private & 0.3181 & 0.0008 & 0.8650 \\
\bottomrule
\end{tabular}
\end{table}

\subsection{Reward-model and PPO Learning-rate Sensitivity}
\label{app:rebuttal_hyperparameters}
Table~\ref{tab:ppo_params} specifies downstream PPO hyperparameters.
The additional reward-model learning-rate search is reported separately in
Table~\ref{tab:rebuttal_rm_lr}. Within the tested grid, the lowest reported loss
occurs at $10^{-4}$ for DP-SGD and RR, $10^{-3}$ for DP-IHT, and $10^{-2}$ for
RR-IHT. 
\begin{table}[htbp]
\centering
\caption{Reward-model learning-rate sweep. Loss is lower-is-better and
preference accuracy (Acc.) is higher-is-better.}
\label{tab:rebuttal_rm_lr}
\begin{tabular}{lcccccc}
\toprule
& \multicolumn{2}{c}{$10^{-2}$} & \multicolumn{2}{c}{$10^{-3}$}
& \multicolumn{2}{c}{$10^{-4}$} \\
\cmidrule(lr){2-3}\cmidrule(lr){4-5}\cmidrule(lr){6-7}
Method & Loss & Acc. & Loss & Acc. & Loss & Acc. \\
\midrule
DP-SGD & 0.5378 & 0.8398 & 0.5328 & 0.8336 & 0.5154 & 0.8302 \\
RR     & 0.5188 & 0.8476 & 0.5247 & 0.8437 & 0.4944 & 0.8568 \\
DP-IHT & 0.3654 & 0.8425 & 0.3316 & 0.8583 & 0.3587 & 0.8554 \\
RR-IHT & 0.3209 & 0.8646 & 0.3384 & 0.8596 & 0.3328 & 0.8621 \\
\bottomrule
\end{tabular}
\end{table}

Table~\ref{tab:rebuttal_ppo_lr} reports the additional PPO learning-rate sweep
for safety alignment with Gemma-3-1B at $\varepsilon=8$. The highest reported
win rates within this grid are $59\%$ for RR-IHT, $57\%$ for DP-IHT, $55\%$ for
RR, and $47\%$ for DP-SGD. 

\begin{table}[htbp]
\centering
\caption{PPO learning-rate sweep for safety alignment with Gemma-3-1B
($\varepsilon=8$), as reported in the rebuttal. Higher win rate is better.
The comparison policy and evaluation sample size remain to be specified.}
\label{tab:rebuttal_ppo_lr}
\begin{tabular}{lccc}
\toprule
Learning rate & $1\times10^{-5}$ & $2\times10^{-5}$ & $4\times10^{-5}$ \\
\midrule
Label-DP RR-IHT & $57\%$ & $57\%$ & $59\%$ \\
Sample-DP IHT   & $55\%$ & $53\%$ & $57\%$ \\
RR              & $51\%$ & $55\%$ & $53\%$ \\
DP-SGD          & $47\%$ & $45\%$ & $41\%$ \\
\bottomrule
\end{tabular}
\end{table}



\vspace{-0.1in}\section{Supported Lemmas and Facts}

\begin{lemma}[Lemma 3.4 in \citet{cai2021cost}]\label{lem:cai_3.4}
Let $S$ and $\{w_i\}_{i=1}^s$ be as defined in Algorithm~\ref{alg:peeling}.  
For any subsets $R_1 \subseteq S$ and $R_2 \subseteq S^{c}$ satisfying $|R_1|=|R_2|$, and for any $c>0$, the following inequality holds:
$$
\|v_{R_2}\|_2^2 \leq(1+c)\|v_{R_1}\|_2^2+4(1+\frac{1}{c}) \sum_{i \in[s]}\|w_i\|_{\infty}^2
$$
where $v\in\mathbb{R}^d$ denotes the input vector to Algorithm~\ref{alg:peeling}.
\end{lemma}

\begin{lemma}[Lemma A. 3 in \citet{cai2021cost}]\label{lem:cai_a3}
    Consider Algorithm~\ref{alg:peeling} with input vector $\widetilde{v}$ and the index set $S$. For any index set $I$, any $v \in \mathbb{R}^{|I|}$ which is a subvector of $\widetilde{v}$ and $\hat{v}$ such that $\|\hat{v}\|_0 \leq \hat{s} \leq s$, we have that for every $c>0$,
    $$
    \|v_S-v\|_2^2 \leq(1+\frac{1}{c}) \frac{|I|-s}{|I|-\hat{s}}\|\hat{v}-v\|_2^2+4(1+c) \sum_{i \in[s]}\|w_i\|_{\infty}^2.
    $$
\end{lemma}

\begin{lemma}[Lemma A.1 in \citet{cai2021cost}]\label{lem:cai_a1}
    For a random vector $w \in \mathbb{R}^d$, where $w_i \sim \operatorname{Lap}(\lambda)$, then for any $\zeta>0$,
$$
\operatorname{Pr}\left(\|w\|_{\infty}^2 \geq 4 \lambda^2 \log ^2 \frac{1}{\zeta} \log ^2 k\right) \leq \zeta .
$$
\end{lemma}

\begin{fact}\label{fact:log}
    Let $\Sigma:=\mathbb{E}[x x^{\top}]$, and suppose : there exist constants $0<\mu \leq \gamma<\infty$ such that $\mu \leq \lambda_{\min }(\Sigma) \leq \lambda_{\max }(\Sigma) \leq \gamma$.
    Given assumptions of covariance and parameters, the following facts hold for the problem setup:
    \begin{itemize}
        \item [(1).] $L_{\mathcal{D}}$ is $\frac{\gamma}{4}$-smooth, that is $\|\nabla L_{\mathcal{D}}(\theta)-\nabla L_{\mathcal{D}}(\theta^{\prime})\|_2 \leq \frac{\gamma}{4}\|\theta-\theta^{\prime}\|_2$.
        \item [(2).]  $L_{\mathcal{D}}$ is $\alpha$-strongly convex on $\{\|\theta\|_2 \leq B\}$ with $\alpha=m(B) \mu=\mu \frac{e^B}{(1+e^B)^2} .$ Thus for all $\theta, \theta^{\prime}$ with $\|\theta\|_2,\|\theta^{\prime}\|_2 \leq B$,
        $
        L_{\mathcal{D}}(\theta^{\prime}) \geq L_{\mathcal{D}}(\theta)+\langle\nabla L_{\mathcal{D}}(\theta), \theta^{\prime}-\theta\rangle+\frac{\alpha}{2}\|\theta^{\prime}-\theta\|_2^2 .
        $
    \end{itemize}
\end{fact}

\begin{fact}\label{fact:ineq1}
For any $c>1$, it holds $(1-\frac{1}{c})\|a\|^2-(c-1)\|b\|_2^2 \leq\|a+b\|^2 \leq(1+\frac{1}{c})\|a\|_2^2+(1+c)\|b\|_2^2$.
\end{fact}

\vspace{-0.1in}\section{Main Proofs}\label{sec:app_proofs}
\vspace{-0.1in}\subsection{Proof of \cref{thm:sample_dp_upper_bound}}
The privacy guarantee relies on the Parallel Composition Theorem \cite{dwork2014algorithmic} (via sample splitting) and the privacy properties of the Peeling Mechanism. Since each iteration operates on disjoint data and satisfies $(\varepsilon, \delta)$-DP via the Peeling Mechanism \cite{cai2021cost}, Algorithm 1 satisfies $(\varepsilon, \delta)$-Sample-DP for the entire dataset.

For simplicity, we denote $\widetilde{g}^t=\frac{1}{m} \sum_{x \in {D}_t} \nabla L(x, w^t), g^t=\nabla L_{\mathcal{D}}(w^t)$, $S^t=\operatorname{supp}(w^t), S^{t+1}=\operatorname{supp}(w^{t+1}), S^*=\operatorname{supp}(w^*)$ and $I^t=S^{t+1} \bigcup S^t \bigcup S^*$. We can see that $|S^t| \leq s,|S^{t+1}| \leq s$ and $|I^t| \leq 2 s+s^*$. We also denote $W^t=4 \sum_{i \in[s]}\|w_i\|_{\infty}^2$, where $\{w_i\}$ are the vectors in Algorithm~\ref{alg:peeling} in the $t$-th iteration. We let $\gamma=\lambda_{\text {max }}(\mathbb{E}[x x^T]), \mu=\lambda_{\text {min }}(\mathbb{E}[x x^T])$ and $\eta_0=\frac{4\eta}{\gamma}$ for some $\eta$.

According to Facts~\ref{fact:log}, we have:
\begin{equation}\label{eq:smooth}
\begin{aligned}
 L_{\mathcal{D}}(w^{t+0.75})-L_{\mathcal{D}}(w^t)  &\leq\langle w^{t+0.75}-w^t, g^t\rangle+\frac{\gamma}{8}\|w^{t+0.75}-w^t\|_2^2 \\
& =\langle w_{I^t}^{t+0.75}-w_{I^t}^t, g_{I^t}^t\rangle+\frac{\gamma}{8}\|w_{I^t}^{t+0.75}-w_{I^t}^t\|_2^2 \\
& \leq \frac{\gamma}{8}\|w_{I^t}^{t+0.75}-w_{I^t}^t+\frac{4\eta}{\gamma} g_{I^t}^t\|_2^2-\frac{2\eta^2}{ \gamma}\|g_{I^t}^t\|_2^2+(1-\eta)\underbrace{\langle w^{t+0.75}-w^t, g^t\rangle}_{\text{Term 3}}.
\end{aligned}
\end{equation}

From the definition of space, it holds that $w^{t+0.75}=\hat{w}^{t+0.75}+\widetilde{w}_{S^{t+1}}$, where $\hat{w}^{t+0.75}=(w^t-\eta_0 \widetilde{g}^t)_{S^{t+1}}$. Therefore, Term 3 can be derived as follows:
\begin{equation}
\begin{aligned}
\text{Term 3} = \langle w^{t+0.75}-w^t, g^t\rangle & =\langle w_{S^{t+1} \cup S^t}^{t+0.75} -w_{S^{t+1}\cup S^t}^t , g_{S^{t+1} \cup S^t }^t \rangle \\
& =\langle w_{S^{t+1}}^{t+0.75}-w_{S^{t+1}}^t, g_{S^{t+1}}^t\rangle+\langle w_{S^t \backslash S^{t+1}}^{t+0.75}-w_{S^t \backslash S^{t+1}}^t, g_{S^t \backslash S^{t+1}}^t\rangle \\
& =\langle w_{S^{t+1}}^{t+0.75}-w_{S^{t+1}}^t, g_{S^{t+1}}^t\rangle-\langle w_{S^t \backslash S^{t+1}}^t, g_{S^t \backslash S^{t+1}}^t\rangle\\
&=\underbrace{\langle\hat{w}_{S^{t+1}}^{t+0.75}-w_{S^{t+1}}^t, g_{S^{t+1}}^t\rangle}_{\text{Term 3.1}}+\langle\widetilde{w}_{S^{t+1}}, g_{S^{t+1}}^t\rangle-\langle w_{S^t \backslash S^{t+1}}^t-g_{S^t \backslash S^{t+1}}^t\rangle .
\end{aligned}
\end{equation}

With the update strategy, we have:

\begin{equation}
\begin{aligned}
\text{Term 3.1} =\langle\hat{w}_{S^{t+1}}^{t+0.75}-w_{S^{t+1}}^t, g_{S^{t+1}}^t\rangle & =\langle-\eta_0 \widetilde{g}_{S^{t+1}}^t, g_{S^{t+1}}^t\rangle=-\frac{4\eta}{\gamma}\langle\widetilde{g}_{S^{t+1}}^t, g_{S^{t+1}}^t\rangle \\
& =-\frac{4\eta}{\gamma}\|g_{S^{t+1}}^t\|_2^2-\frac{4\eta}{\gamma}\langle\widetilde{g}_{S^{t+1}}^t-g_{S^{t+1}}^t, g_{S^{t+1}}^t\rangle \\
& \leq-\frac{4\eta}{\gamma}\|g_{S^{t+1}}^t\|_2^2+\frac{2\eta}{ \gamma}\|g_{S^{t+1}}^t\|_2^2+\frac{2\eta}{ \gamma}\|\widetilde{g}_{S^{t+1}}^t-g_{S^{t+1}}^t\|_2^2 \\
& =-\frac{2\eta}{ \gamma}\|g_{S^{t+1}}^t\|_2^2+\frac{2\eta}{ \gamma}\|\widetilde{g}_{S^{t+1}}^t-g_{S^{t+1}}^t\|_2^2.
\end{aligned}
\end{equation}

Substituting the above into Term 3, we obtain, for any $c \textgreater 1$:
\begin{equation}
\text{Term 3} \leq-\frac{2\eta}{ \gamma}\|g_{S^{t+1}}^t\|_2^2+\frac{2\eta}{ \gamma}\|\widetilde{g}_{S^{t+1}}^t-g_{S^{t+1}}^t\|_2^2+c\|\widetilde{w}_{S^{t+1}}\|_2^2+\frac{1}{4 c}\|g_{S^{t+1}}^t\|_2^2 \underbrace{-\langle w_{S^t \backslash S^{t+1}}^t-g_{S^t \backslash S^{t+1}}^t\rangle}_{\text{Term 3.2}} .
\end{equation}

Then, it holds that:
\begin{equation}
\begin{aligned}
\text{Term 3.2} = -\langle w_{S^t \backslash S^{t+1}}^t-g_{S^t \backslash S^{t+1}}^t\rangle & \leq \frac{\gamma}{8\eta}(\|w_{S^t \backslash S^{t+1}}^t-\frac{4\eta}{\gamma} g_{S^t \backslash S^{t+1}}^t\|_2^2-(\frac{4\eta}{\gamma})^2\|g_{S^t \backslash S^{t+1}}^t\|_2^2) \\
& =\frac{\gamma}{8 \eta}\|w_{S^t \backslash S^{t+1}}^t-\frac{4\eta}{\gamma} g_{S^t \backslash S^{t+1}}^t\|_2^2-\frac{2\eta}{ \gamma}\|g_{S^t \backslash S^{t+1}}^t\|_2^2.
\end{aligned}
\end{equation}

Let $v=w^t-\frac{4\eta}{\gamma} \widetilde{g}^t, R_2=S^t \backslash S^{t+1}$ and $R_1=S^{t+1} \backslash S^t$. From Lemma~\ref{lem:cai_3.4}, for $c>1$, we have:
\begin{equation}
\|w_{S^t \backslash S^{t+1}}^t-\frac{4\eta}{\gamma} \widetilde{g}_{S^t \backslash S^{t+1}}^t\|_2^2 \leq(1+\frac{1}{c})\|w_{S^{t+1} \backslash S^t}^t-\frac{4\eta}{\gamma} \widetilde{g}_{S^{t+1} \backslash S^t}^t\|_2^2+(1+c) W^t .
\end{equation}

Due to Fact~\ref{fact:ineq1}, it also holds:
\begin{equation}\label{eq:fact2}
\begin{aligned}
& (1-\frac{1}{c})\|w_{S^t \backslash S^{t+1}}^t-\frac{4\eta}{\gamma} g_{S^t \backslash S^{t+1}}^t\|_2^2-(c-1) \frac{16\eta^2}{\gamma^2}\|g_{S^t \backslash S^{t+1}}^t-\widetilde{g}_{S^t \backslash S^{t+1}}^t\|_2^2 \\
& \leq\|w_{S^t \backslash S^{t+1}}^t-\frac{\eta}{4\gamma} \widetilde{g}_{S^t \backslash S^{t+1}}^t\|_2^2 \leq(1+\frac{1}{c})\|w_{S^{t+1} \backslash S^t}^t-\frac{4\eta}{\gamma} \widetilde{g}_{S^{t+1} \backslash S^t}^t\|_2^2+(1+c) W^t \\
& \leq(1+\frac{1}{c})[(1+1 / c)\|w_{S^{t+1} \backslash S^t}^t-\frac{4\eta}{\gamma} g_{S^{t+1} \backslash S^t}^t\|_2^2+(1+c) \frac{16\eta^2}{\gamma^2}\|g_{S^t \backslash S^{t+1}}^t-\widetilde{g}_{S^t \backslash S^{t+1}}^t\|_2^2]+2(1+c) W^t.
\end{aligned}
\end{equation}

That is,
\begin{equation}
\begin{aligned}
& \|w_{S^t \backslash S^{t+1}}^t-\frac{4\eta}{\gamma} g_{S^t \backslash S^{t+1}}^t\|_2^2 \leq \frac{(c+1)^2}{c(c-1)}\|w_{S^{t+1} \backslash S^t}^t-\frac{4 \eta}{\gamma} g_{S^{t+1} \backslash S^t}^t\|_2^2 \\
& \quad+(c+\frac{(c+1)^2}{c}) \frac{4\eta}{2 \gamma}(\|g_{S^t \backslash S^{t+1}}^t-\widetilde{g}_{S^t \backslash S^{t+1}}^t\|_2^2+\|g_{S^{t+1} \backslash S^t}^t-\widetilde{g}_{S^{t+1} \backslash S^t}^t\|_2^2)+\frac{c(1+c)}{c-1} W^t.
\end{aligned}
\end{equation}

Plugging the above into Term 3.2 yields:
\begin{equation}
\begin{aligned}
& -\langle w_{S^t \backslash S^{t+1}}^t-g_{S^t \backslash S^{t+1}}^t\rangle \leq \frac{(c+1)^2}{2 c(c-1)} \frac{4\eta}{\gamma}\|g_{S^{t+1} \backslash S^t}^t\|_2^2 \\
+ & \frac{\gamma c(1+c)}{8 \eta(c-1)} W^t-\frac{2\eta}{ \gamma}\|g_{S^t \backslash S^{t+1}}^t\|_2^2+\frac{(2 c+3) 2\eta}{ \gamma}(\|g_{S^t \backslash S^{t+1}}^t-\widetilde{g}_{S^t \backslash S^{t+1}}^t\|_2^2+\|g_{S^{t+1} \backslash S^t}^t-\widetilde{g}_{S^{t+1} \backslash S^t}^t\|_2^2) .
\end{aligned}
\end{equation}

Therefore, for Term 3, we have:
\begin{equation}
\begin{aligned}
\text{Term 3} &= \langle w^{t+0.75}-w^t, g^t\rangle \\
&\leq-\frac{2\eta}{ \gamma}\|g_{S^{t+1}}^t\|_2^2+\frac{2\eta}{ \gamma}\|\widetilde{g}_{S^{t+1}}^t-g_{S^{t+1}}^t\|_2^2+c\|\widetilde{w}_{S^{t+1}}\|_2^2 +\frac{1}{4 c}\|g_{S^{t+1}}^t\|_2^2-\langle w_{S^t \backslash S^{t+1}}^t-g_{S^t \backslash S^{t+1}}^t\rangle \\
& \leq-\frac{2\eta}{ \gamma}\|g_{S^{t+1}}^t\|_2^2+\frac{2\eta}{ \gamma}\|\widetilde{g}_{S^{t+1}}^t-g_{S^{t+1}}^t\|_2^2+c\|\widetilde{w}_{S^{t+1}}\|_2^2+\frac{1}{4 c}\|g_{S^{t+1}}^t\|_2^2-\frac{2\eta}{ \gamma}\|g_{S^{t+1}}^t\|_2^2 \\
+& \frac{(c+1)^2}{ c(c-1)} \frac{2\eta}{\gamma}\|g_{S^{t+1} \backslash S^t}^t\|_2^2+\frac{(2 c+3) 2\eta}{ \gamma}(\|g_{S^t \backslash S^{t+1}}^t-\widetilde{g}_{S^t \backslash S^{t+1}}^t\|_2^2+\|g_{S^{t+1} \backslash S^t}^t-\widetilde{g}_{S^{t+1} \backslash S^t}^t\|_2^2) \\
&= \frac{2\eta}{ \gamma}\|g_{S^{t+1} \backslash S^t}^t\|_2^2+\frac{2\eta}{ \gamma} \frac{3 c+1}{c(c-1)}\|g_{S^{t+1} \backslash S^t}^t\|_2^2-\frac{2\eta}{ \gamma}\|g_{S^t \backslash S^{t+1}}^t\|_2^2-\frac{2\eta}{ \gamma}\|g_{S^{t+1}}^t\|_2^2 \\
+&\frac{1}{4 c}\|g_{S^{t+1}}^t\|_2^2+\frac{\gamma c(1+c)}{8 \eta(c-1)} W^t+\frac{2\eta}{ \gamma}\|\widetilde{g}_{S^{t+1}}^t-g_{S^{t+1}}^t\|_2^2+c\|\widetilde{w}_{S^{t+1}}\|_2^2 \\
+ &\frac{(2 c+3) \eta}{4\gamma}(\|g_{S^t \backslash S^{t+1}}^t-\widetilde{g}_{S^t \backslash S^{t+1}}^t\|_2^2+\|g_{S^{t+1} \backslash S^t}^t-\widetilde{g}_{S^{t+1} \backslash S^t}^t\|_2^2).
\end{aligned}
\end{equation}

Denote $N^t = \frac{\gamma c(1+c)}{8 \eta(c-1)} W^t+\frac{2\eta}{ \gamma}\|\widetilde{g}_{S^{t+1}}^t-g_{S^{t+1}}^t\|_2^2+c\|\widetilde{w}_{S^{t+1}}\|_2^2+\frac{(2 c+3) 4\eta}{\gamma}(\|g_{S^t \backslash S^{t+1}}^t-\widetilde{g}_{S^t \backslash S^{t+1}}^t\|_2^2+\|g_{S^{t+1} \backslash S^t}^t-\widetilde{g}_{S^{t+1} \backslash S^t}^t\|_2^2)$:
\begin{equation}
\begin{aligned}
\text{Term 3} &= \langle w^{t+0.75}-w^t, g^t\rangle \leq \frac{2\eta}{ \gamma}\|g_{S^{t+1} \backslash S^t}^t\|_2^2-\frac{2\eta}{ \gamma}\|g_{S^t \backslash S^{t+1}}^t\|_2^2-\frac{2\eta}{ \gamma}\|g_{S^{t+1}}^t\|_2^2+C_1 \frac{4\eta}{\gamma c}\|g_{S^{t+1}}^t\|_2^2+N^t,
\end{aligned}
\end{equation}
where $C_1>0$ is some constant. It can be observed that:
\begin{equation}
    \begin{aligned}
    \frac{2\eta}{ \gamma}\|g_{S^{t+1} \backslash S^t}^t\|_2^2-\frac{2\eta}{ \gamma}\|g_{S^t \backslash S^{t+1}}^t\|_2^2-\frac{2\eta}{ \gamma}\|g_{S^{t+1}}^t\|_2^2 & =-\frac{2\eta}{ \gamma}\|g_{S^t \backslash S^{t+1}}^t\|_2^2-\frac{2\eta}{ \gamma}\|g_{S^{t+1} \cap S^t}^t\|_2^2 \\
    & =-\frac{2\eta}{ \gamma}\|g_{S^{t+1} \cup S^t}^t\|_2^2.
    \end{aligned}
\end{equation}

In summary, it holds that
\begin{equation}
\begin{aligned}
\text{Term 3} &= \langle w^{t+0.75}-w^t, g^t\rangle \leq -\frac{2\eta}{ \gamma}\|g_{S^{t+1} \cup S^t}^t\|_2^2+C_1 \frac{4\eta}{\gamma c}\|g_{S^{t+1}}^t\|_2^2+N^t.
\end{aligned}
\end{equation}

Recall Eq.~\eqref{eq:smooth}, we have
\begin{equation}
\begin{aligned}\label{eq:loss_term1}
 L_{\mathcal{D}}(w^{t+0.75})-L_{\mathcal{D}}(w^t) &\leq \frac{\gamma}{8}\|w_{I^t}^{t+0.75}-w_{I^t}^t+\frac{4\eta}{\gamma} g_{I^t}^t\|_2^2-\frac{2\eta^2}{ \gamma}\|g_{I^t}^t\|_2^2+(1-\eta) * \text{Term 3} \\
& \leq \frac{\gamma}{8}\|w_{I^t}^{t+0.75}-w_{I^t}^t+\frac{4\eta}{\gamma} g_{I^t}^t\|_2^2-\frac{2\eta^2}{ \gamma}\|g_{I^t}^t\|_2^2 \\
-&\frac{(1-\eta) 2\eta}{ \gamma}\|g_{S^{t+1} \cup S^t}^t\|_2^2+C_1 \frac{(1-\eta)}{c} \frac{4\eta}{\gamma}\|g_{S^{t+1}}^t\|_2^2 +(1-\eta) N^t \\
& \leq \frac{\gamma}{8}\|w_{I^t}^{t+0.75}-w_{I^t}^t+\frac{4\eta}{\gamma} g_{I^t}^t\|_2^2-\frac{2\eta^2}{ \gamma}\|g_{I^t \backslash(S^t \cup S^*)}^t\|_2^2-\frac{2\eta^2}{ \gamma}\|g_{(S^t \cup S^*)}^t\|_2^2 \\
-&\frac{(1-\eta) 2\eta}{ \gamma}\|g_{S^{t+1} \cup S^t}^t\|_2^2  +C_1 \frac{(1-\eta)}{c} \frac{4\eta}{\gamma}\|g_{S^{t+1}}^t\|_2^2+(1-\eta) N^t \\
& \leq \underbrace{\frac{\gamma}{8}\|w_{I^t}^{t+0.75}-w_{I^t}^t+\frac{4\eta}{\gamma} g_{I^t}^t\|_2^2-\frac{2\eta^2}{ \gamma}\|g_{I^t \backslash(S^t \cup S^*)}^t\|_2^2}_{\text{Term 1}}-\frac{2\eta^2}{ \gamma}\|g_{(S^t \cup S^*)}^t\|_2^2 \\
-&\frac{(1-\eta) 2\eta}{ \gamma}\|g_{S^{t+1} \backslash(S^* \cup S^t)}^t\|_2^2+C_1 \frac{(1-\eta)}{c} \frac{4\eta}{\gamma}\|g_{S^{t+1}}^t\|_2^2+(1-\eta) N^t,
\end{aligned}
\end{equation}
where the last inequality is due to $S^{t+1} \backslash(S^* \bigcup S^t) \subseteq S^{t+1} \bigcup S^t$.

We now focus on Term 1. Choose $R \subseteq S^t \backslash S^{t+1}$ such that $|R|=|I^t \backslash(S^* \cup S^t)|=\mid S^{t+1} \backslash(S^t \cup S^*) \mid$. Taking $v=w^t-\frac{4\eta}{\gamma} \widetilde{g}^t, R_2=R$, and $R_1=I^t \backslash(S^* \cup S^t)$ in Lemma~\ref{lem:cai_3.4} yields, for $c>1$,
\begin{equation}
\|w_R^t-\frac{4\eta}{\gamma} \widetilde{g}_R^t\|_2^2 \leq(1+c)\|(w^t-\frac{4\eta}{\gamma} \widetilde{g}^t)_{I^t \backslash(S^* \cup S^t)}\|_2^2+(1+\frac{1}{c}) W^t .
\end{equation}

According to Fact~\ref{fact:ineq1} and Eq.\eqref{eq:fact2}, we have:
\begin{equation}\label{eq:fact_R}
(\frac{4\eta}{\gamma})^2\|g_{I^t \backslash(S^* \cup S^t)}^t\|_2^2 \geq(1-\frac{1}{c})\|w_R^t-\frac{4\eta}{\gamma} g_R^t\|_2^2-c W^t-c \frac{16\eta^2}{\gamma^2}(\|\widetilde{g}_R^t-g_R^t\|_2^2+\|g_{I^t \backslash(S^* \cup S^t)}^t-\widetilde{g}_{I^t \backslash(S^* \cup S^t)}^t\|_2^2) .
\end{equation}

Then, for Term 1, it holds that:
\begin{equation}
\begin{aligned}
\text{Term 1} & =  \frac{\gamma}{8}\|w_{I^t}^{t+0.75}-w_{I^t}^t+\frac{4\eta}{\gamma} g_{I^t}^t\|_2^2-\frac{2\eta^2}{ \gamma}\|g_{I^t \backslash(S^t \cup S^*)}^t\|_2^2 \\
& \leq \frac{\gamma}{8}\|\widetilde{w}_{S^{t+1}}\|_2^2+\frac{\gamma}{8}\|\hat{w}_{I^t}^{t+0.75}-w_{I^t}^t+\frac{4\eta}{\gamma} g_{I^t}^t\|_2^2-\frac{\gamma}{8}(1-\frac{1}{c})\|w_R^t-\frac{4\eta}{\gamma} g_R^t\|_2^2+\frac{\gamma c}{2} W^t \\
& +c \frac{2\eta^2}{ \gamma}(\|\widetilde{g}_R^t-g_R^t\|_2^2+\|g_{I^t \backslash(S^* \cup S^t)}^t-\widetilde{g}_{I^t \backslash(S^* \cup S^t)}^t\|_2^2) \\
& =\frac{\gamma}{8}\|\hat{w}_{I^t}^{t+0.75}-w_{I^t}^t+\frac{4\eta}{\gamma} g_{I^t}^t\|_2^2-\frac{\gamma}{8}\|\hat{w}_R^{t+0.75}-w_R^t+\frac{4\eta}{\gamma} g_R^t\|_2^2+\frac{\gamma}{8}\|\widetilde{w}_{S^{t+1}}\|_2^2+\frac{\gamma c}{2} W^t \\
& +\frac{\gamma}{2 c}\|w_R^t-\frac{4\eta}{\gamma} g_R^t\|_2^2+c \frac{2\eta^2}{ \gamma}(\|\widetilde{g}_R^t-g_R^t\|_2^2+\|g_{I^t \backslash(S^* \cup S^t)}^t-\widetilde{g}_{I^t \backslash(S^* \cup S^t)}^t\|_2^2),\\
& \leq \frac{\gamma}{8}\|\hat{w}_{I^t \backslash R}^{t+0.75}-w_{I^t \backslash R}^t+\frac{4\eta}{\gamma} g_{I^t \backslash R}^t\|_2^2+\frac{\gamma}{2 c}(1+\frac{1}{c})\|\frac{4\eta}{\gamma} g_{I^t \backslash(S^* \cup S^t)}^t\|_2^2+\frac{\gamma}{8}\|\widetilde{w}_{S^{t+1}}\|_2^2 \\
& +\underbrace{\frac{\gamma c}{8} W^t+\frac{\gamma}{8 c}(1+c) W^t+C_2 c \frac{8\eta^2}{ \gamma}(\|\widetilde{g}_R^t-g_R^t\|_2^2+\|g_{I^t \backslash(S^* \cup S^t)}^t-\widetilde{g}_{I^t \backslash(S^* \cup S^t)}^t\|_2^2)}_{N_1^t}
\end{aligned}
\end{equation}
where the first inequality comes from $[\hat{w}_{I^t}^{t+0.75}-(w_{I^t}^t-\frac{4\eta}{\gamma} g_{I^t}^t)]_{S^{t+1}}=0$, thus $\langle\widetilde{w}_{S^{t+1}}, \hat{w}_{I^t}^{t+0.75}-(w_{I^t}^t-\frac{4\eta}{\gamma} g_{I^t}^t)\rangle=0$ and the second equality is due to $\hat{w}_R^{t+0.75}=0$.

Denote $N_1^t = \frac{\gamma c}{8} W^t+\frac{\gamma}{8 c}(1+c) W^t+C_2 c \frac{8\eta^2}{ \gamma}(\|\widetilde{g}_R^t-g_R^t\|_2^2+\|g_{I^t \backslash(S^* \cup S^t)}^t-\widetilde{g}_{I^t \backslash(S^* \cup S^t)}^t\|_2^2)$, with Fact~\ref{fact:ineq1}, Eq.\eqref{eq:fact_R} and $w_{I^t \backslash(S^* \cup S^t)}^t=0$, then by combining the above, the Term 1 follows:
\begin{equation}
\begin{aligned}
\text{Term 1} & \leq \frac{\gamma}{8}\underbrace{\|\hat{w}_{I^t \backslash R}^{t+0.75}-w_{I^t \backslash R}^t+\frac{4\eta}{\gamma} g_{I^t \backslash R}^t\|_2^2}_{\text{Term 1.1}}+\frac{\gamma}{2 c}(1+\frac{1}{c})\|\frac{4\eta}{\gamma} g_{I^t \backslash(S^* \cup S^t)}^t\|_2^2+\frac{\gamma}{8}\|\widetilde{w}_{S^{t+1}}\|_2^2 + N_1^t.
\end{aligned}
\end{equation}
In Lemma~\ref{lem:cai_a3}, qith $v=w_{I^t \backslash R}^t-\frac{4\eta}{\gamma} \widetilde{g}_{I^t \backslash R}^t, \hat{v}=w^*, S=S^{t+1}$, for all $c>1$, we have: 
\begin{equation}
    \text{Term 1.1} = \|\hat{w}_{I^t \backslash R}^{t+0.75}-w_{I^t \backslash R}^t+\frac{4\eta}{\gamma} \widetilde{g}_{I^t \backslash R}^t\|_2^2 \leq(1+\frac{1}{c}) \frac{|I^t \backslash R|-s}{|I^t \backslash R|-s^*}\|w^*-w_{I^t \backslash R}^t+\frac{4\eta}{\gamma} \widetilde{g}_{I^t \backslash R}^t\|_2^2+(1+c) W^t .
\end{equation}

With Fact~\ref{fact:ineq1}:
\begin{equation}
\begin{aligned}
& (1-\frac{1}{c})\|\hat{w}_{I^t \backslash R}^{t+0.75}-w_{I^t \backslash R}^t+\frac{4\eta}{\gamma} g_{I^t \backslash R}^t\|_2^2-(c-1) \frac{\eta^2}{\gamma^2}\|g_{I^t \backslash R}^t-\widetilde{g}_{I^t \backslash R}^t\|_2^2 \\
& \leq\|\hat{w}_{I^t \backslash R}^{t+0.75}-w_{I^t \backslash R}^t+\frac{4\eta}{\gamma} \widetilde{g}_{I^t \backslash R}^t\|_2^2 \\
& \leq(1+\frac{1}{c}) \frac{|I^t \backslash R|-s}{|I^t \backslash R|-s^*}\|w^*-w_{I^t \backslash R}^t+\frac{4\eta}{\gamma} \widetilde{g}_{I^t \backslash R}^t\|_2^2+(1+c) W^t \\
& \leq(1+\frac{1}{c}) \frac{|I^t \backslash R|-s}{|I^t \backslash R|-s^*}[(1+\frac{1}{c})\|w^*-w_{I^t \backslash R}^t+\frac{4\eta}{\gamma} g_{I^t \backslash R}^t\|_2^2+(1+c) \frac{16\eta^2}{\gamma^2}\|g_{I^t \backslash R}^t-\widetilde{g}_{I^t \backslash R}^t\|_2^2]+(1+c) W^t.
\end{aligned}
\end{equation}

That is 
\begin{equation}
\begin{aligned}
\text{Term 1.1} &
= \| \hat{w}_{I^t \backslash R}^{t+0.75}-w_{I^t \backslash R}^t+  \frac{4\eta}{\gamma} g_{I^t \backslash R}^t\|_2^2 \\
&\leq \frac{(c+1)^2}{c(c-1)} \frac{2 s^*}{s+s^*}\| w^*-w_{I^t \backslash R}^t+\frac{4\eta}{\gamma} g_{I^t \backslash R}^t \|_2^2 \\
& +\frac{(c+1)^2}{c-1} \frac{16\eta^2}{\gamma^2}\|g_{I^t \backslash R}^t-\widetilde{g}_{I^t \backslash R}^t\|_2^2+c \frac{16\eta^2}{\gamma^2}\|g_{I^t \backslash R}^t-\widetilde{g}_{I^t \backslash R}^t\|_2^2+\frac{(1+c) c}{c-1} W^t .
\end{aligned}
\end{equation}

Let $c \geq \frac{\sqrt{3}}{\sqrt{3}-\sqrt{2}}$ and denote $N_3^t = C_3 c(\|g_{I^t \backslash R}^t-\widetilde{g}_{I^t \backslash R}^t\|_2^2+\|g_{I^t \backslash R}^t-\widetilde{g}_{I^t \backslash R}^t\|_2^2+W^t)$, we have the following due to $|I^t \backslash R| \leq 2 s^*+s$:
\begin{equation}
\|\hat{w}_{I^t \backslash R}^{t+0.75}-w_{I^t \backslash R}^t+\frac{4\eta}{\gamma} g_{I^t \backslash R}^t\|_2^2 \leq \frac{3}{2} \frac{2 s^*}{s+s^*}\|w^*-w_{I^t \backslash R}^t+\frac{4\eta}{\gamma} g_{I^t \backslash R}^t\|_2^2 + N_3^t
\end{equation}

Substituting the above to Term 1, it holds that:
\begin{equation}
\begin{aligned}
\text{Term 1} & =  \frac{\gamma}{8}\|w_{I^t}^{t+0.75}-w_{I^t}^t+\frac{4\eta}{\gamma} g_{I^t}^t\|_2^2-\frac{2\eta^2}{ \gamma}\|g_{I^t \backslash(S^t \cup S^*)}^t\|_2^2 \\
& \leq \frac{3 \gamma s^*}{8(s+s^*)}\|w^*-w_{I^t \backslash R}^t+\frac{4\eta}{\gamma} g_{I^t \backslash R}^t\|_2^2+\frac{\gamma}{8 c}(1+\frac{1}{c})\|\frac{4\eta}{\gamma} g_{I^t \backslash(S^* \cup S^t)}^t\|_2^2+\frac{\gamma}{8}\|\widetilde{w}_{S^{t+1}}\|_2^2 +N_1^t+N_3^t \\
& =\frac{3 \gamma s^*}{2(s+s^*)}\|w^*-w_{I^t \backslash R}^t+\frac{4\eta}{\gamma} g_{I^t \backslash R}^t\|_2^2+\frac{\gamma}{8 c}(1+\frac{1}{c})\|\frac{4\eta}{\gamma} g_{S^{t+1}}^t\|_2^2 +\frac{\gamma}{8}\|\widetilde{w}_{S^{t+1}}\|_2^2+N_1^t+N_3^t \\
& =\frac{3 s^*}{s+s^*}(\eta\langle w^*-w^t, g^t\rangle+\frac{\gamma}{8}\|w^*-w^t\|_2^2+\frac{2\eta^2}{ c \gamma}\|g_{I^t}^t\|_2^2)+\frac{2\eta^2}{ c \gamma}(1+\frac{1}{c})\|g_{S^{t+1}}^t\|_2^2 +\frac{\gamma}{8}\|\widetilde{w}_{S^{t+1}}\|_2^2+N_1^t+N_3^t \\
& \leq \frac{3 s^*}{s+s^*}(\eta(L_{\mathcal{D}}(w^*)-L_{\mathcal{D}}(w^t))+\frac{\gamma/4-\eta \mu}{2}\|w^*-w^t\|_2^2+\frac{9\eta^2}{ c \gamma}\|g_{I^t}^t\|_2^2)+\frac{\eta^2}{8 c \gamma}(1+\frac{1}{c})\|g_{S^{t+1}}^t\|_2^2 \\
& \quad+\underbrace{\frac{\gamma}{8}\|\widetilde{w}_{S^{t+1}}\|_2^2+N_1^t+N_3^t}_{N_2^t}.
\end{aligned}
\end{equation}

Combining above to Eq.\eqref{eq:loss_term1}, we have:
\begin{equation}
\begin{aligned}
& L_{\mathcal{D}}\left(w^{t+0.75}\right)-L_{\mathcal{D}}\left(w^t\right) \\
&\leq \frac{\gamma}{8}\left\|w_{I^t}^{t+0.75}-w_{I^t}^t+\frac{4\eta}{\gamma} g_{I^t}^t\right\|_2^2-\frac{2\eta^2}{ \gamma}\left\|g_{I^t \backslash\left(S^t \cup S^*\right)}^t\right\|_2^2-\frac{2\eta^2}{ \gamma}\left\|g_{\left(S^t \cup S^*\right)}^t\right\|_2^2 \\
& -\frac{2(1-\eta) \eta}{ \gamma}\left\|g_{S^{t+1} \backslash\left(S^* \cup S^t\right)}^t\right\|_2^2+C_1 \frac{(1-\eta)}{c} \frac{4\eta}{\gamma}\left\|g_{S^{t+1}}^t\right\|_2^2+(1-\eta) N^t \\
& \leq \frac{3 s^*}{s+s^*}\left(\eta\left(L_{\mathcal{D}}\left(w^*\right)-L_{\mathcal{D}}\left(w^t\right)\right)+\frac{\gamma/4-\eta \mu}{2}\left\|w^*-w^t\right\|_2^2+\frac{2\eta^2}{ c \gamma}\left\|g_{I^t}^t\right\|_2^2\right)+\frac{2\eta^2}{ c \gamma}\left(1+\frac{1}{c}\right)\left\|g_{S^{t+1}}^t\right\|_2^2 \\
& -\frac{2\eta^2}{ \gamma}\left\|g_{\left(S^t \cup S^*\right)}^t\right\|_2^2-\frac{2(1-\eta) \eta}{ \gamma}\left\|g_{S^{t+1} \backslash\left(S^* \cup S^t\right)}^t\right\|_2^2+C_1 \frac{(1-\eta)}{c} \frac{4\eta}{\gamma}\left\|g_{S^{t+1}}^t\right\|_2^2\left\|g_{S^{t+1}}^t\right\|_2^2+(1-\eta) N^t+N_2^t.
\end{aligned}
\end{equation}
When $c \rightarrow \infty$, we have $\frac{\eta^2}{2 c \gamma}\left(1+\frac{1}{c}\right)\left\|g_{S^{t+1}}^t\right\|_2^2+C_1 \frac{(1-\eta)}{c} \frac{4\eta}{\gamma}\left\|g_{S^{t+1}}^t\right\|_2^2 \rightarrow 0$.
Thus, if $\eta \geq \frac{1}{2}$ there must exits a sufficient large $c$ such that
\begin{equation}
\begin{aligned}
& \frac{2\eta^2}{ c \gamma}\left(1+\frac{1}{c}\right)\left\|g_{S^{t+1}}^t\right\|_2^2+\frac{(1-\eta)}{c}\left(\frac{1}{4}+\frac{2\eta}{ \gamma}\right)\left\|g_{S^{t+1}}^t\right\|_2^2 \leq \frac{\eta(1-\eta)}{ \gamma}\left\|g_{S^{t+1}}^t\right\|_2^2 \\
& \leq \frac{\eta^2}{ \gamma}\left\|g_{\left(S^t \cup S^*\right)}^t\right\|_2^2+\frac{(1-\eta) \eta}{ \gamma}\left\|g_{S^{t+1} \backslash\left(S^* \cup S^t\right)}^t\right\|_2^2.
\end{aligned}
\end{equation}

Therefore,
\begin{equation}
\begin{aligned}
& L_{\mathcal{D}}\left(w^{t+0.75}\right)-L_{\mathcal{D}}\left(w^t\right) \\
\leq& \frac{\gamma}{8}\left\|w_{I^t}^{t+0.75}-w_{I^t}^t+\frac{4\eta}{\gamma} g_{I^t}^t\right\|_2^2-\frac{2\eta^2}{ \gamma}\left\|g_{I^t \backslash\left(S^t \cup S^*\right)}^t\right\|_2^2-\frac{2\eta^2}{ \gamma}\left\|g_{\left(S^t \cup S^*\right)}^t\right\|_2^2 \\
& -\frac{(1-\eta) 2\eta}{ \gamma}\left\|g_{S^{t+1} \backslash\left(S^* \cup S^t\right)}^t\right\|_2^2+C_1 \frac{(1-\eta)}{c} \frac{4\eta}{\gamma}\left\|g_{S^{t+1}}^t\right\|_2^2+(1-\eta) N^t \\
\leq &  \frac{3 s^*}{s+s^*}\left(\eta\left(L_{\mathcal{D}}\left(w^*\right)-L_{\mathcal{D}}\left(w^t\right)\right)+\frac{\gamma/4-\eta \mu}{2}\left\|w^*-w^t\right\|_2^2+\frac{2\eta^2}{ c \gamma}\left\|g_{I^t}^t\right\|_2^2\right) \\
& -\frac{\eta^2}{ \gamma}\left\|g_{\left(S^t \cup S^*\right)}^t\right\|_2^2-\frac{(1-\eta) \eta}{ \gamma}\left\|g_{S^{t+1} \backslash\left(S^* \cup S^t\right)}^t\right\|_2^2+(1-\eta) N^t+N_2^t.
\end{aligned}
\end{equation}

Take $\eta=\frac{2}{3}, s=6 \frac{\gamma^2}{\mu^2} s^*$ so that $\frac{3 s^*}{s+s^*} \leq \frac{\mu^2}{6 \gamma(\gamma/4-\eta \mu)} \leq \frac{1}{8}$. We have
\begin{equation}
\begin{aligned}
& L_{\mathcal{D}}\left(w^{t+0.75}\right)-L_{\mathcal{D}}\left(w^t\right)\\
\leq& \frac{3 s^*}{s+s^*}\left(\eta\left(L_{\mathcal{D}}\left(w^*\right)-L_{\mathcal{D}}\left(w^t\right)\right)+\frac{\gamma/4-\eta \mu}{2}\left\|w^*-w^t\right\|_2^2+\frac{2\eta^2}{ c \gamma}\left\|g_{I^t}^t\right\|_2^2\right) \\
& -\frac{\eta^2}{ \gamma}\left\|g_{\left(S^t \cup S^*\right)}^t\right\|_2^2-\frac{(1-\eta) \eta}{ \gamma}\left\|g_{S^{t+1} \backslash\left(S^* \cup S^t\right)}^t\right\|_2^2+(1-\eta) N^t+N_2^t \\
\leq&  \frac{2 s^*}{s+s^*}\left(L_{\mathcal{D}}\left(w^*\right)-L_{\mathcal{D}}\left(w^t\right)\right)+\frac{\mu^2}{412 \gamma}\left\|w^*-w^t\right\|_2^2+\frac{4}{9 \gamma}\left\|g_{I^t}^t\right\|_2^2 \\
&-\frac{4}{9 \gamma}\left\|g_{\left(S^t \cup S^*\right)}^t\right\|_2^2-\frac{2}{9 \gamma}\left\|g_{S^{t+1} \backslash\left(S^* \cup S^t\right)}^t\right\|_2^2+O\left(N^t+N_2^t\right) \\
\leq &  \frac{2 s^*}{s+s^*}\left(L_{\mathcal{D}}\left(w^*\right)-L_{\mathcal{D}}\left(w^t\right)\right)-\frac{3}{9 \gamma}\left(\left\|g_{\left(S^t \cup S^*\right)}^t\right\|_2^2-\frac{\mu^2}{4}\left\|w^*-w^t\right\|_2^2\right)+O\left(N^t+N_2^t\right) \\
\leq & \left(\frac{2 s^*}{s+s^*}+\frac{\mu}{6 \gamma}\right)\left(L_{\mathcal{D}}\left(w^*\right)-L_{\mathcal{D}}\left(w^t\right)\right)+O\left(N^t+N_2^t\right),
\end{aligned}
\end{equation}
where the penultimate inequality is due to the following Lemma~\ref{lem:jain_5}.
\begin{lemma}[Lemma 5 in \citet{jain2014iterative}]\label{lem:jain_5}
$$
\| g_{\left(S^t \cup S^*\right)}^t\|_2^2-\frac{\mu^2}{4}\| w^*-w^t \|_2^2 \geq \frac{\mu}{2}\left(L_{\mathcal{D}}\left(w^t\right)-L_{\mathcal{D}}\left(w^*\right)\right).
$$
\end{lemma}

Therefore, we have:
\begin{equation}
L_{\mathcal{D}}\left(w^{t+0.75}\right)-L_{\mathcal{D}}\left(w^*\right) \leq\left(1-\frac{5}{18} \frac{\mu}{\gamma}\right)\left(L_{\mathcal{D}}\left(w^t\right)-L_{\mathcal{D}}\left(w^*\right)\right)+O\left(N^t+N_2^t\right),
\end{equation}
where 
\begin{equation}
\begin{aligned}
    &N^t+N_2^t  \leq N^t + \frac{\gamma}{8}\left\|\widetilde{w}_{S^{t+1}}\right\|_2^2+N_1^t+N_3^t \\
    & \leq \frac{\gamma c(1+c)}{8 \eta(c-1)} W^t+\frac{2\eta}{ \gamma}\|\widetilde{g}_{S^{t+1}}^t-g_{S^{t+1}}^t\|_2^2+c\|\widetilde{w}_{S^{t+1}}\|_2^2+\frac{(2 c+3) 4\eta}{\gamma}(\|g_{S^t \backslash S^{t+1}}^t-\widetilde{g}_{S^t \backslash S^{t+1}}^t\|_2^2+\|g_{S^{t+1} \backslash S^t}^t-\widetilde{g}_{S^{t+1} \backslash S^t}^t\|_2^2) \\
    &+ \frac{\gamma}{8}\left\|\widetilde{w}_{S^{t+1}}\right\|_2^2\\
    &+  \frac{\gamma c}{8} W^t+\frac{\gamma}{8 c}(1+c) W^t+C_2 c \frac{8\eta^2}{ \gamma}(\|\widetilde{g}_R^t-g_R^t\|_2^2+\|g_{I^t \backslash(S^* \cup S^t)}^t-\widetilde{g}_{I^t \backslash(S^* \cup S^t)}^t\|_2^2)\\
    &+C_3 c(\|g_{I^t \backslash R}^t-\widetilde{g}_{I^t \backslash R}^t\|_2^2+\|g_{I^t \backslash R}^t-\widetilde{g}_{I^t \backslash R}^t\|_2^2+W^t)\\
    &\leq O\left(\sum\left\|w_i\right\|_{\infty}^2+\left\|\widetilde{g}^t-g^t\right\|_{2}^2+s\|\widetilde{w}\|_{\infty}^2\right)\\
    &\leq O\left(\sum\left\|w_i\right\|_{\infty}^2+\left(2 s+s^*\right)\left\|\widetilde{g}^t-g^t\right\|_{\infty}^2+s\|\widetilde{w}\|_{\infty}^2\right),
\end{aligned}
\end{equation}
where each coordinate of $w_i, \widetilde{w}$ sampled from $\sim \operatorname{Lap}\left(O\left(\frac{ \eta_0  T \sqrt{s \log \frac{1}{\delta}}}{n \varepsilon}\right)\right)$. 
We first bound the term of $\sum_{i \in[s]}\left\|w_i\right\|_{\infty}^2+s\|\widetilde{w}\|_{\infty}^2$. Recall Lemma~\ref{lem:cai_a1}, with probability at least $1-\zeta$, we take the union:
\begin{equation}
    \sum_{i \in[s]}\left\|w_i\right\|_{\infty}^2+s\|\widetilde{w}\|_{\infty}^2 \leq O\left(\frac{ \eta_0^2 s^2 T^2 \log ^2 \frac{s}{\zeta} \log \frac{1}{\delta}}{n^2 \varepsilon^2}\right)
\end{equation}
Therefore,
\begin{equation}
    N^t+N_2^t \leq O\left(\frac{ \eta_0^2 s^2 T^2 \log ^2 \frac{s}{\zeta} \log \frac{1}{\delta}}{n^2 \varepsilon^2}\right) + O\left(\frac{sT\log \frac{1}{\delta}}{n}\right)
\end{equation}
According to Fact~\ref{fact:log}, we have:
\begin{equation}
\frac{\gamma}{8}\left\|w-w^*\right\|_2^2 \geq L_{\mathcal{D}}(w)-L_{\mathcal{D}}\left(w^*\right) \geq \frac{m(B) \mu}{2}\left\|w-w^*\right\|_2^2 .
\end{equation}
Therefore,
\begin{equation}\label{eq:up_induc}
L_{\mathcal{D}}\left(w^{t+0.75}\right)-L_{\mathcal{D}}\left(w^*\right) \leq\left(1-\frac{5}{18} \frac{\mu}{\gamma}\right)\left(L_{\mathcal{D}}\left(w^t\right)-L_{\mathcal{D}}\left(w^*\right)\right)+O\left(\frac{ \eta_0^2 s^2 T^2 \log ^2 \frac{s}{\zeta} \log \frac{1}{\delta}}{n^2 \varepsilon^2}\right) + O\left(\frac{sT\log \frac{1}{\delta}}{n}\right).
\end{equation}
In what follows, we show that $w^{t+1}=w^{t+0.75}$ for all $t$. We proceed by induction.
Assume that $w^{i+1}=w^{i+0.75}$ holds for all $i \in\{0,1, \ldots, t-1\}$. We will show that it also holds for $i=t$.
Applying Eq.\eqref{eq:up_induc} for each $i \in\{0,1, \ldots, t-1\}$, we have:
\begin{equation}\label{eq:up_l_w}
\begin{aligned}
 &\frac{m(B) \mu}{2}\left\|w^{t+0.75}-w^*\right\|_2^2 \leq L_{\mathcal{D}}\left(w^{t+0.75}\right)-L_{\mathcal{D}}\left(w^*\right) \\
& \leq\left(1-\frac{5}{18} \frac{\mu}{\gamma}\right)^t\left(L_{\mathcal{D}}\left(w^1\right)-L_{\mathcal{D}}\left(w^*\right)\right)+ O\left(\frac{\gamma}{m(B)\mu}\frac{ \eta_0^2 s^2 T^2 \log ^2 \frac{s}{\zeta} \log \frac{1}{\delta}}{n^2 \varepsilon^2}\right) + O\left(\frac{\gamma}{m(B)\mu}\frac{ sT\log \frac{1}{\delta}}{n}\right) \\
& \leq \gamma\left(1-\frac{5}{18} \frac{\mu}{\gamma}\right)^t\left\|w^1-w^*\right\|_2^2+O\left(\frac{\gamma}{m(B)\mu}\frac{ \eta_0^2 s^2 T^2 \log ^2 \frac{s}{\zeta} \log \frac{1}{\delta}}{n^2 \varepsilon^2}\right) + O\left(\frac{\gamma}{m(B)\mu}\frac{sT\log \frac{1}{\delta}}{n}\right).
\end{aligned}
\end{equation}
When $\left\|w^1-w^*\right\|_2^2 \leq \frac{B}{2} \frac{m(B)\mu}{\gamma}$, and $n$ is large enough such that
$$
n^2 \geq O\left(\frac{\gamma}{\mu m(B)} \frac{s^2 T^2 \log ^2 \frac{s}{\zeta} \log \frac{1}{\delta}}{ \varepsilon^2}\right).
$$
Then, it holds that $\left\|w^{t+0.75}-w^*\right\|_2^2 \lesssim B$ , thus $w^{t+1}=w^{t+0.75}$, which concludes
\begin{equation}
L_{\mathcal{D}}\left(w^{T+1}\right)-L_{\mathcal{D}}\left(w^*\right) \leq\left(1-\frac{5}{18} \frac{\mu}{\gamma}\right)^T\left(L_{\mathcal{D}}\left(w^t\right)-L_{\mathcal{D}}\left(w^*\right)\right)+O\left(\frac{\gamma}{\mu} \frac{ \eta_0^2 s^2 T^2 \log ^2 \frac{s}{\zeta} \log \frac{1}{\delta}}{n^2 \varepsilon^2}\right) + O\left(\frac{\gamma}{\mu} \frac{sT\log \frac{1}{\delta}}{n}\right).
\end{equation}
Thus, take $T=\widetilde{O}\left(\frac{\gamma}{\mu} \log n\right)$ and Eq.\eqref{eq:up_l_w}, we complete the proof.

\vspace{-0.1in}\subsection{Proof of \cref{thm:sample_dp_lower_bound}}
Before presenting the proof, let us introduce some useful definitions and lemmas. Score attack introduced by \cite{cai2023score} refers to a form of tracing attack applicable to general parametric families of distributions. Formally, let $\mathcal{P}=\left\{f_\theta: \theta \in \Theta\right\}$ be a statistical model with a well-defined score statistic $S(x ; \theta)=\nabla_\theta \log f_\theta(x)$. Given a private estimator $\hat{\theta}=M(X)$ computed on a dataset $X$, and a candidate data point $x$, the Score Attack statistic $A_\theta(x, \hat{\theta})$ is defined as the correlation between the estimation error and the score function at the candidate point:
$$
A_\theta(x, \hat{\theta})=\langle\hat{\theta}-\theta, S(x ; \theta)\rangle.
$$
Since DP requires these two estimators to be indistinguishable, the success of the Score Attack implies a violation of privacy. Therefore, to satisfy DP, the estimator cannot be too accurate, which leads to the following:
\begin{lemma}[Theorem 2.1 in \citet{cai2023score}]
    Let $X=\left\{x_1, x_2, \cdots, x_n\right\}$ be an i.i.d. sample drawn from $f_{{\theta}}$. For each $i \in [n]$, let $X_i^{\prime}$ denote an adjacent data set of $X$ obtained by replacing $x_i$ with an independent copy $x_i^{\prime} \sim f_{{\theta}}$, then it holds that:
    \begin{itemize}
        \item [1.] Soundness: for each $i \in[n]$,
        $$
        \mathbb{E} A_{{\theta}}\left(x_i, M\left(X_i^{\prime}\right)\right)=0 ; \mathbb{E}\left|A_{{\theta}}\left(x_i, M\left(X_i^{\prime}\right)\right)\right| \leq \sqrt{\mathbb{E}\|M(X)-{\theta}\|_2^2} \sqrt{\lambda_{\max }({I}({\theta}))},
        $$
        where ${I}({\theta})$ is the Fisher information matrix of $f_{{\theta}}$.
        \item [2.] Completeness: if for every $j \in[d], \log f_{{\theta}}(X)$ is continuously differentiable with respect to $\theta_j$ and $\left|\frac{\partial}{\partial \theta_j} \log f_{{\theta}}(X)\right|<g_j(X)$ such that $\mathbb{E}\left|g_j(X) M(X)_j\right|<\infty$, we have  
        $$
        \sum_{i \in[n]} \mathbb{E} A_{{\theta}}\left(x_i, M(X)\right)=\sum_{j \in[d]} \frac{\partial}{\partial \theta_j} \mathbb{E} M(X)_j .
        $$
    \end{itemize}
\end{lemma}
Our lower bounds leverage the above score attack. Specifically, considering i.i.d. observations $\left(y_1, x_1\right), \cdots,\left(y_n, x_n\right)$ drawn from $f_{\theta^{\star}}\left(y_i \mid x_i\right)=\exp \left(y_i\left\langle\theta^{\star}, x_i\right\rangle-\psi\left(\left\langle\theta^{\star}, x_i\right\rangle\right)\right); \quad x_i \sim \mathcal{D}_z$ with $\|\theta\|_0 \leq s^*$, where $y_i \in \{0, 1\}$ is the preference label (mapped from $\{-1, +1\}$ labels to the canonical form), $\psi(u) = \log(1 + e^u)$ is the log-partition function (cumulant function) for the Bernoulli distribution and the dispersion term $c(\sigma)$ from the original Eq.(5.1) in \citet{cai2023score} is $1$ for logistic models. Therefore, we have the following lemma:
\begin{lemma}[Proposition 5.1 in \citet{cai2023score}]\label{lem:cai_prop51}
    Suppose $\Sigma$ is diagonal and according to assumptions that $\lambda_{\max }(\Sigma) \leq \gamma,\|x\|_{\inf}<1$, and $\left\|\psi^{\prime \prime}\right\|_{\infty}<1/4$. If the estimator $M$ is $(\varepsilon, \delta)$-differentially private with $0<\varepsilon<1$ and satisfies $\|M(y, X)-\theta\|_2^2 \leq R \lesssim  s^*$, then
    $$
    \sum_{i \in[n]} \mathbb{E} A_i \leq  n \varepsilon \sqrt{\mathbb{E}\|M(y, X)-\theta\|_2^2} \sqrt{\gamma}+ 2 \sqrt{2} \delta s^* \sqrt{ \log (1 / \delta)} .
    $$
\end{lemma}

\begin{lemma}[Proposition 5.2 in \citet{cai2023score}]\label{lem:cai_prop52}
    Suppose $s^* \lesssim \sqrt{d}$, and $s^* \gg  \log ^2\left(d / s^*\right)$. For every $M$ satisfying $\mathbb{E}_{y, X \mid \theta}\|M(y, x)-\theta\|_2^2 \lesssim 1$ at every $\theta$, we have
    $$
    \sum_{i \in[n]} \mathbb{E}_{\pi} \mathbb{E}_{y, X \mid \theta} A_i \gtrsim s^* \log \left(d / s^*\right)
    $$
    where $\pi$ refers to the sparse truncated normal prior for $\theta$.
\end{lemma}
Based on above preliminaries, we are ready to prove \cref{thm:sample_dp_lower_bound}. Consider the parameter space $\Theta=\left\{\theta \in \mathbb{R}^d:\|\theta\|_0 \leq s^*,\|\theta\|_{\infty} \leq 1\right\}$. It is clear that assumptions of \cref{thm:sample_dp_lower_bound} are sufficient to apply Lemma~\ref{lem:cai_prop51} and Lemma~\ref{lem:cai_prop52}. Therefore, we have:
\begin{equation}
s^* \log d \lesssim \sum_{i \in[n]} \mathbb{E}_{{\pi}} \mathbb{E}_{{y}, {X} \mid \theta} A_i \leq  n \varepsilon \sqrt{\mathbb{E}\|M({y}, {X})-\theta\|_2^2} \sqrt{\gamma }+ 2\sqrt{2} \delta s^* \sqrt{ \log (1 / \delta) },
\end{equation}
which indicates 
\begin{equation}
 n \varepsilon \mathbb{E}_{{\pi}} \sqrt{\mathbb{E}_{{y}, {X} \mid \theta}\|M({y}, {X})-\theta\|_2^2} \sqrt{\gamma} \gtrsim s^* \log \left(d / s^*\right)- 2 \sqrt{2} n \delta s^* \sqrt{ \log (1 / \delta) / c(\sigma)} .
\end{equation}
Also, we have $\delta<n^{-(1+c_3)}$ for some $c_3>0$ implies that for $n$ sufficiently large, $s^* \log \left(d / s^*\right)- 2 \sqrt{2} n \delta s^* \sqrt{\log (1 / \delta)} \gtrsim s^* \log \left(d / s^*\right)$. We then conclude that
\begin{equation}
    \sup_{\theta \in \Theta} \mathbb{E}_{{\pi}} \mathbb{E}_{{y}, {X} \mid {\theta}}\|M({y}, {X})-{\theta}\|_2^2 \gtrsim \mathbb{E}_{{\pi}} \mathbb{E}_{{y}, {X} \mid {\theta}}\|M({y}, {X})-{\theta}\|_2^2 \gtrsim \frac{\left(s^* \log \left(d / s^*\right)\right)^2}{\gamma n^2 \varepsilon^2} .
\end{equation}
The above bound holds for any $M$ satisfying $\mathbb{E}_{y, X \mid \theta}\|M(y, x)-\theta\|_2^2 \lesssim 1$, according to assumption that $s^* \log(d/s^*) \lesssim n \varepsilon$, then we can extend the results to all $M \in \mathcal{M}_{\varepsilon,\delta}$. Combining with the non-private minimax lower bound, then the proof is completed.

\vspace{-0.1in}\subsection{Proof of \cref{thm:label_dp_upper_bound}}
To verify the privacy guarantee, consider any two adjacent inputs (labels) $y, y' \in \{0, 1\}$ (where $y \neq y'$) and any output $\tilde{y} \in \{0, 1\}$. The likelihood ratio is bounded as follows:$$\frac{\mathbb{P}(\mathcal{M}_{RR}(y) = \tilde{y})}{\mathbb{P}(\mathcal{M}_{RR}(y') = \tilde{y})} \le \frac{\max \mathbb{P}}{\min \mathbb{P}} = \frac{\frac{e^\varepsilon}{e^\varepsilon + 1}}{\frac{1}{e^\varepsilon + 1}} = e^\varepsilon.$$Thus, the generation of each $\tilde{y}_i$ satisfies $\varepsilon$-DP. Since the labels are processed independently, the full privatized label set $\tilde{D} = \{(x_i, \tilde{y}_i)\}_{i=1}^n$ satisfies $\varepsilon$-Label-DP. 
According to the Post-Processing Property of differential privacy, any computation performed on the output of a differentially private mechanism maintains the original privacy guarantee. Therefore, the final output $\theta^T$ satisfies $\varepsilon$-Label-DP. 

In the following, we will prove the utility. We denote $g^t = \nabla L_{\mathcal{D}}(w^{t-1})$ and $\widetilde{g}^t= \frac{e^{\varepsilon}+1}{e^{\varepsilon}-1} \cdot (\hat{g}^t -\frac{\nabla L_{x \in D}(w^{t-1},x,y=0)}{e^{\varepsilon}+1} - \frac{\nabla L_{x \in D} (w^{t-1},x,y=1)}{e^{\varepsilon}+1} )$ as the debiased gradient, where $\hat{g}^t = \nabla L_{x \in D} (w^{t-1},x,y=\hat{y})$ is noisy gradient after random response mechanism. Then, the proof of utility is almost the same as the proof of \cref{thm:sample_dp_upper_bound} without noise addition in the index selection and final output. Following the steps, we have:
\begin{equation}
L_{\mathcal{D}}\left(w^{t+0.75}\right)-L_{\mathcal{D}}\left(w^*\right) \leq\left(1-\frac{5}{18} \frac{\mu}{\gamma}\right)\left(L_{\mathcal{D}}\left(w^t\right)-L_{\mathcal{D}}\left(w^*\right)\right)+O\left(N^t+N_2^t\right),
\end{equation}
where 
\begin{equation}
\begin{aligned}
    &N^t+N_2^t  \leq N^t + \frac{\gamma}{8}\left\|\widetilde{w}_{S^{t+1}}\right\|_2^2+N_1^t+N_3^t \\
    & \leq \frac{\gamma c(1+c)}{8 \eta(c-1)} W^t+\frac{2\eta}{ \gamma}\|\widetilde{g}_{S^{t+1}}^t-g_{S^{t+1}}^t\|_2^2+c\|\widetilde{w}_{S^{t+1}}\|_2^2+\frac{(2 c+3) 4\eta}{\gamma}(\|g_{S^t \backslash S^{t+1}}^t-\widetilde{g}_{S^t \backslash S^{t+1}}^t\|_2^2+\|g_{S^{t+1} \backslash S^t}^t-\widetilde{g}_{S^{t+1} \backslash S^t}^t\|_2^2) \\
    &+ \frac{\gamma}{8}\left\|\widetilde{w}_{S^{t+1}}\right\|_2^2\\
    &+  \frac{\gamma c}{8} W^t+\frac{\gamma}{8 c}(1+c) W^t+C_2 c \frac{8\eta^2}{ \gamma}(\|\widetilde{g}_R^t-g_R^t\|_2^2+\|g_{I^t \backslash(S^* \cup S^t)}^t-\widetilde{g}_{I^t \backslash(S^* \cup S^t)}^t\|_2^2)\\
    &+C_3 c(\|g_{I^t \backslash R}^t-\widetilde{g}_{I^t \backslash R}^t\|_2^2+\|g_{I^t \backslash R}^t-\widetilde{g}_{I^t \backslash R}^t\|_2^2+W^t)\\
    &\leq O\left(\sum\left\|w_i\right\|_{\infty}^2+\left\|\widetilde{g}^t-g^t\right\|_{2}^2+s\|\widetilde{w}\|_{\infty}^2\right)\\
    &\leq O\left(\sum\left\|w_i\right\|_{\infty}^2+\left(2 s+s^*\right)\left\|\widetilde{g}^t-g^t\right\|_{\infty}^2+s\|\widetilde{w}\|_{\infty}^2\right).
\end{aligned}
\end{equation}
One difference from \cref{thm:sample_dp_upper_bound} is that the first and last terms are zero, because no noise is injected during index selection or in the final released output. Another difference is that $\widetilde{g}^t$ denotes the debiased gradient after RR mechanisms instead of $\bar{g}^t = \frac{1}{m} \sum_{x \in D_t} \nabla L\left(x, w^t\right)$. Thus, we have with probability at least $1-\delta$, for all $t \in[T]$,
\begin{equation}
\left\|\widetilde{g}^t-g^t\right\|_{2}^2 \leq \left\|\widetilde{g}^t-\bar{g}^t\right\|_{2}^2 + \left\|\bar{g}^t - {g}^t \right \|_{2}^2\leq O\left(\frac{s\log \frac{1}{\delta}}{n} + \frac{s\log \frac{1}{\delta}}{n \varepsilon^2} \right)
\end{equation}
Therefore, 
\begin{equation}
L_{\mathcal{D}}\left(w^{t+0.75}\right)-L_{\mathcal{D}}\left(w^*\right) \leq\left(1-\frac{5}{18} \frac{\mu}{\gamma}\right)\left(L_{\mathcal{D}}\left(w^t\right)-L_{\mathcal{D}}\left(w^*\right)\right)+O\left(\frac{s\log \frac{1}{\delta}}{n} + \frac{s\log \frac{1}{\delta}}{n \varepsilon^2} \right).
\end{equation}
When $\left\|w^1-w^*\right\|_2^2 \leq \frac{B}{2} \frac{m(B)\mu}{\gamma}$, and $n$ is large enough such that
$$
n \geq O\left(\frac{\gamma}{\mu m(B)} \frac{s \log \log \frac{1}{\delta}}{ \varepsilon^2}\right).
$$
Then, it holds that the update still in the ball, which concludes.

\vspace{-0.1in}\subsection{Proof of \cref{thm:label_dp_lower_bound}}
For the private part, we consider the set of all binary vectors in $\{0, 1\}^d$ with a Hamming weight of exactly $s$. By the Gilbert-Varshamov bound, there exists a subset of these vectors, denoted by $\Omega = \{ \omega_1, \dots, \omega_M \}$, such that:
\begin{itemize}
    \item Cardinality: $\log M \ge \frac{s}{8} \log\left(\frac{d}{s}\right)$.
    \item Sparsity: $\|\omega_j\|_0 = s$ for all $j \in [M]$.
    \item Hamming Separation: For any distinct pair $i \ne j$, the Hamming distance satisfies $\|\omega_i - \omega_j\|_H \ge \frac{s}{2}$.
\end{itemize}
We define our parameter set $\mathcal{V}=\left\{\theta_1, \ldots, \theta_M\right\}$ by scaling the vectors in $\Omega$ such that $\theta_j=\tau \cdot \omega_j$. Thus, we have:
\begin{equation}
    \|\theta_i - \theta_j\|_2^2 = \tau^2 \|\omega_i - \omega_j\|_H \ge \tau^2 \frac{s}{2}.
    \label{eq:theta_omega}
\end{equation}

Let $y_t \in \{0, 1\}$ be the true label and $\widetilde{y}_t \in \{0, 1\}$ be the private label released by Randomized Response, then we have:
\begin{equation}
    \mathbb{P}(\widetilde{y}_t = 1 \mid y_t = 1) = \frac{e^\varepsilon}{1+e^\varepsilon}, \qquad \mathbb{P}(\widetilde{y}_t = 1 \mid y_t = 0) = \frac{1}{1+e^\varepsilon}.
\end{equation}
Let $\eta_t(\theta) = \sigma(\langle z_t, \theta \rangle)$ be the probability that the true label $y_t=1$. The probability that the private label $\widetilde{y}_t=1$, denoted by $q_t(\theta)$, is:
\begin{equation}
    q_t(\theta) = \eta_t(\theta) \cdot \frac{e^\varepsilon}{1+e^\varepsilon} + (1-\eta_t(\theta)) \cdot \frac{1}{1+e^\varepsilon}= C_\varepsilon \cdot \eta_t(\theta) + \frac{1}{1+e^\varepsilon},
\end{equation}
where $C_\varepsilon = \frac{e^\varepsilon - 1}{e^\varepsilon + 1}$. For Randomized Response, it is clear that the output probability $q_t(\theta)$ is always bounded within $[\frac{1}{1+e^\varepsilon}, \frac{e^\varepsilon}{1+e^\varepsilon}]$, which indicates that $\frac{1}{q_t(1-q_t)} \le 4e$.

Let $Q_{\theta}^n$ denote the joint distribution of the private labels $\widetilde{y}_1, \dots, \widetilde{y}_n$ generated by parameter $\theta$, then it holds that
\begin{equation}
    \begin{aligned}
        &D_{KL}(Q_{\theta_i}^n \| Q_{\theta_j}^n) = \sum_{t=1}^n D_{KL}(Q_{\theta_i, t} \| Q_{\theta_j, t}) \overset{(i)}{\le} \sum_{t=1}^n 4e \left( q_t(\theta_i) - q_t(\theta_j) \right)^2 = K_1 C_\varepsilon^2 \left( \eta_t(\theta_i) - \eta_t(\theta_j) \right)^2\\
        & \leq \sum_{t=1}^n K_1 C_\varepsilon^2 \left( \eta_t(\theta_i) - \eta_t(\theta_j) \right)^2 = \sum_{t=1}^n K_1 C_\varepsilon^2 \left( \sigma(\langle z_t, \theta_i \rangle) - \sigma(\langle z_t, \theta_j \rangle) \right)^2 \le \sum_{t=1}^n \frac{K_1 C_\varepsilon^2}{16} \langle z_t, \theta_i - \theta_j \rangle^2\\
        & \le C \cdot \gamma n \varepsilon^2 s \tau^2,
    \end{aligned}
\end{equation} 
where $C$ is the constant.

According DP Fano's inequality, we also have $C \gamma n \varepsilon^2 s \tau^2 \le \frac{1}{2} \cdot \frac{s}{8} \log\left(\frac{d}{s}\right)$, which indicates that $\tau \le c' \sqrt{\frac{\log(d/s)}{n \gamma \varepsilon^2}}.$
Therefore, we have
\begin{equation}
    \mathbb{E}[\|\hat{\theta} - \theta^\star\|_2] \ge \Omega\left( \sqrt{s} \cdot \sqrt{\frac{\log(d/s)}{n \gamma \varepsilon^2}} \right) = \Omega\left( \sqrt{\frac{s \log(d/s)}{n \gamma \varepsilon^2}} \right).
\end{equation}

For the non-private part, we consider the same construction and similar proof ideas, and the only difference is the KL distance. Specifically, Let $P_{\theta}^n$ denote the joint distribution of the labels $y_1, \dots, y_n$ given inputs $z_1, \dots, z_n$ and parameter $\theta$. The KL divergence can be bounded by:
\begin{equation}
    \begin{aligned}
        D_{KL}(P_{\theta_i}^n \| P_{\theta_j}^n) = \sum_{t=1}^n D_{KL}(P_{\theta_i, t} \| P_{\theta_j, t}) \le \frac{1}{4} \sum_{t=1}^n \langle z_t, \theta_i - \theta_j \rangle^2 \le \frac{n}{4}  \gamma \|\theta_i - \theta_j\|_2^2 \leq \frac{ns \gamma \tau^2}{2},
    \end{aligned}
\end{equation}
where the penultimate inequality is due to property of the sigmoid function and the last inequality comes from the construction. Thus, considering Fano's inequality, we also have $ \tau \le c' \sqrt{\frac{\log(d/s)}{n \gamma}}$, which indicates that $\mathbb{E}[\|\hat{\theta} - \theta\|_2] \gtrsim \tau\sqrt{s} \asymp \sqrt{s} \cdot \sqrt{\frac{\log(d/s)}{n \gamma }} = \sqrt{\frac{s \log(d/s)}{n \gamma}}$. Thus, combing the above the private result, the proof is completed.

\vspace{-0.1in}\subsection{Proof of Theorem~\ref{thm:general_label_dp_lower_bound}}
We first introduce several auxiliary lemmas. 

\begin{lemma}[Assouad's lemma \cite{chowdhury2024differentially}]\label{lem:Assouad}
   Let $\mathcal{V} \subseteq \mathcal{P}$ be a set of distributions indexed by the hypercube $\mathcal{E}_d=\{ \pm 1\}^d$. Suppose there exists a $\tau \in \mathbb{R}$ and $\alpha>0$, such that $\rho$ satisfies: (i) for all $u, v, w \in \mathcal{E}_d, \rho\left(\theta\left(P_u\right), \theta\left(P_v\right)\right) \geq 2 \tau \cdot \sum_{i=1}^d \mathbb{1}\left(u_i \neq v_i\right)$ and (ii) $\rho\left(\theta\left(P_u\right), \theta\left(P_v\right)\right) \leq \alpha\left(\rho\left(\theta\left(P_u\right), \theta\left(P_w\right)\right)+\rho\left(\theta\left(P_v\right), \theta\left(P_w\right)\right)\right.$ ), i.e., $\alpha$-triangle inequality. For each $i \in[d]$, define the mixture distributions:
$$
P_{+i}:=\frac{2}{\left|\mathcal{E}_d\right|} \sum_{e \in \mathcal{E}_d: e_i=1} P_e \text { and } P_{-i}:=\frac{2}{\left|\mathcal{E}_d\right|} \sum_{e \in \mathcal{E}_d: e_i=-1} P_e .
$$
Then, we have
$$
R(\mathcal{P}, \rho) \geq \frac{\tau}{2 \alpha} \sum_{i=1}^d\left(1-\left\|P_{+i}-P_{-i}\right\|_{\mathrm{TV}}\right) .
$$
\end{lemma}
\begin{lemma}[Assouad's Corollary \cite{chowdhury2024differentially}]\label{lem:Assouad-Corollary}
Under the same conditions of Lemma~\ref{lem:Assouad}, we have
$$
R(\mathcal{P}, \rho) \geq \frac{d \tau}{2 \alpha}\left[1-\left(\frac{1}{d} \sum_{i=1}^d \frac{1}{2^d} \sum_{e \in \mathcal{E}_d}\left\|P_e-P_{\bar{e}^i}\right\|_{\mathrm{TV}}^2\right)^{1 / 2}\right],
$$
where $\bar{e}^i$ is a vector in $\mathcal{E}_d$ that flips the $i$-th coordinate of $e$.
\end{lemma}

\begin{lemma}[Assouad's lemma for central DP \cite{chowdhury2024differentially}]\label{lem:Assouad-central-dp}
    Let the same conditions of Lemma B. 4 hold. If for all $i \in[d]$, there exists a coupling $(X, Y)$ between $P_{+i}$ and $P_{-i}$ with $\mathbb{E}\left[d_{\operatorname{Ham}}(X, Y)\right] \leq D$ for some $D \geq 0$, then
    $$
    R_c(\mathcal{P}, \rho, \varepsilon, \delta) \geq \frac{d \tau}{2 \alpha} \cdot\left(0.9 e^{-10 \varepsilon D}-10 D \delta\right).
    $$
\end{lemma}

Then, we are ready to prove Theorem~\ref{thm:general_label_dp_lower_bound}.

Consider $\Theta_{B,s^*}= \{\theta\in\mathbb R^d:\|\theta\|_0\le s^*,\ \|\theta\|_2\le B\}.$ Fix any support set $S\subset[d]$ with $|S|=s^*$, for example $S=\{1,\ldots,s^*\}$, and define the restricted parameter class $\Theta_S := \{\theta\in\Theta_{B,s^*}: \operatorname{supp}(\theta)\subseteq S\}.$ Let $\mathcal M_{(\varepsilon,\delta)}$ denote the class of all central $(\varepsilon,\delta)$-Label-DP estimators. For an estimator $\mathcal M$, let
$R(\mathcal M,\theta):= \mathbb E_{\theta}\|\mathcal M(D)-\theta\|_2^2 .$ Since $\Theta_S\subseteq \Theta_{B,s^*}$, for every estimator $\mathcal M$,
$\sup_{\theta\in\Theta_{B,s^*}} R(\mathcal M,\theta) \ge \sup_{\theta\in\Theta_S} R(\mathcal M,\theta).$
Taking the infimum over the same algorithm class $\mathcal A_{\rm all}$ on both sides gives
\begin{equation}
\label{eq:submodel-reduction}
    \inf_{\mathcal M\in\mathcal M_{(\varepsilon,\delta)}}
    \sup_{\theta\in\Theta_{B,s^*}} R(\mathcal M,\theta)
    \ge
    \inf_{\mathcal M\in\mathcal M_{(\varepsilon,\delta)}}
    \sup_{\theta\in\Theta_S} R(\mathcal M,\theta).
\end{equation}
Therefore, it is sufficient to prove a lower bound on the smaller class $\Theta_S$. 
Then, we prove the lower bound by reducing sparse BTL reward estimation to testing over the above sparse hypercube.

\paragraph{Sparse hypercube construction.}
Let $m=s^*.$ For every sign vector $v=(v_1,\ldots,v_m)\in\{-1,+1\}^m,$ define
$\theta_v = a\sum_{j=1}^m v_j e_j = (a v_1,\ldots,a v_m,0,\ldots,0),$ where $a>0$ will be chosen later. Then $\|\theta_v\|_0=m=s^*, \|\theta_v\|_2=a\sqrt m.$ Thus, whenever $a\sqrt m\le B,$
we have $\theta_v\in\Theta_S\subseteq\Theta_{B,s^*}.$

For any two vertices $u,v\in\{-1,+1\}^m$, $\|\theta_u-\theta_v\|_2^2 = \left\| a\sum_{j=1}^m (u_j-v_j)e_j \right\|_2^2  = a^2\sum_{j=1}^m (u_j-v_j)^2.$ If $u_j=v_j$, then $(u_j-v_j)^2=0$; if $u_j\neq v_j$, then $(u_j-v_j)^2=4$. Hence $\|\theta_u-\theta_v\|_2^2 = 4a^2 d_{\rm H}(u,v),$
where $d_{\rm H}$ denotes Hamming distance.

Let $v^{(j)}$ denote the vector obtained by flipping the $j$-th coordinate of
$v$. Then $d_{\rm H}(v,v^{(j)})=1,$
and therefore $ \|\theta_v-\theta_{v^{(j)}}\|_2^2=4a^2.$

We condition on the public covariates $X=(x_1,\ldots,x_n).$ Under Label-DP, the covariates are public, and only the labels are private. Therefore, all arguments below can be regarded as conditional on $X$.

For each $v\in\{-1,+1\}^m$, let $P_v$ be the conditional distribution of the
label vector $Y=(y_1,\ldots,y_n)$
under parameter $\theta_v$. Since the BTL model gives
$ y_i\mid x_i \sim {\rm Bernoulli} \bigl(\sigma(\langle x_i,\theta_v\rangle)\bigr),$
and the samples are conditionally independent given $X$, we have $P_v(Y\mid X) = \prod_{i=1}^n P_{\theta_v,i}(y_i\mid x_i),$
where $P_{\theta_v,i} = {\rm Bernoulli} \bigl(\sigma(\langle x_i,\theta_v\rangle)\bigr).$

\paragraph{Non-private term.}
We first apply the classical Assouad lemma to obtain the ordinary statistical lower bound. In the notation of the classical Assouad lemma~\ref{lem:Assouad}, we take $\mathcal E_m=\{-1,+1\}^m, \rho(\theta,\theta')=\|\theta-\theta'\|_2^2.$
The squared Euclidean loss satisfies the $2$-triangle inequality: $\rho(\theta_1,\theta_2) \le 2\rho(\theta_1,\theta_3)+2\rho(\theta_3,\theta_2).$
Thus, in the notation of Assouad's lemma, $\alpha=2.$
Moreover, for any $u,v\in\{-1,+1\}^m$, $\rho(\theta_u,\theta_v) = \|\theta_u-\theta_v\|_2^2= 4a^2 d_{\rm H}(u,v).$ We take $\tau=2a^2$ to satisfy
$\rho(\theta_u,\theta_v) \ge 2\tau d_{\rm H}(u,v).$ 

Now fix $v$ and its one-coordinate flip $v^{(j)}$. For sample $i$, define
$$
\alpha_i=\langle x_i,\theta_v\rangle,
\qquad
\beta_i=\langle x_i,\theta_{v^{(j)}}\rangle.
$$
Since $v^{(j)}$ differs from $v$ only in coordinate $j$, $\theta_v-\theta_{v^{(j)}} = 2a v_j e_j.$
Therefore, $\alpha_i-\beta_i = \langle x_i,\theta_v-\theta_{v^{(j)}}\rangle = 2a v_j x_{ij}.$

Also, we have
$$
{\rm KL}(P_{\theta_v,i}\|P_{\theta_{v^{(j)}},i})+
{\rm KL}(P_{\theta_{v^{(j)}},i}\|P_{\theta_v,i}) \le
(\alpha_i-\beta_i)^2 = 4a^2x_{ij}^2.
$$
Since $P_v$ and $P_{v^{(j)}}$ are product distributions over
$i=1,\ldots,n$, we have:
$$
{\rm KL}(P_v\|P_{v^{(j)}}) + {\rm KL}(P_{v^{(j)}}\|P_v) \le 4a^2\sum_{i=1}^n x_{ij}^2 \leq  4a^2 n\gamma,
$$
where the last inequality comes from the Assumption~\ref{asm:boundness}.
By Pinsker's inequality,
$$
\|P_v-P_{v^{(j)}}\|_{\rm TV}^2 \le C a^2 n\gamma.
$$
Substituting this into Lemma~\ref{lem:Assouad-Corollary}, with $d$ replaced
by $m=s^*$, $\alpha=2$, and $\tau=2a^2$, gives
$$
\inf_{\widehat\theta}
\sup_{v\in\{-1,+1\}^m}
\mathbb E_v\|\widehat\theta-\theta_v\|_2^2
\ge
\frac{m\tau}{2\alpha}
\left(1-C\sqrt{a^2 n\gamma}\right) =
\frac{m a^2}{2}
\left(1-Ca\sqrt{n\gamma}\right).
$$
Choosing $a^2=\frac{c_0}{\gamma n}$ with $c_0>0$ sufficiently small ensures $1-Ca\sqrt{n\gamma}\ge c$
for a universal constant $c>0$. Therefore,
\[
    \inf_{\widehat\theta}
    \sup_{v\in\{-1,+1\}^m}
    \mathbb E_v\|\widehat\theta-\theta_v\|_2^2
    \gtrsim
    m a^2
    =
    \frac{s^*}{\gamma n}.
\]
Since $\mathcal M_{(\varepsilon,\delta)}$ is a subset of all estimators, the same lower bound holds for estimators satisfying central Label-DP:
$$
\inf_{\mathcal M\in \mathcal M_{(\varepsilon,\delta)}}
    \sup_{\theta\in\Theta_{B,s^*}}
    \mathbb E_\theta\|\mathcal M(D)-\theta\|_2^2
    \gtrsim
    \frac{s^*}{\gamma n}.
$$

\paragraph{Central Label-DP private term.}
For each coordinate $j\in[m]$, define the mixture distributions
$$
P_{+j} = \frac{1}{2^{m-1}} \sum_{v:v_j=+1} P_v,
\qquad
P_{-j} = \frac{1}{2^{m-1}} \sum_{v:v_j=-1} P_v.
$$
By the central-DP Assouad lemma~\ref{lem:Assouad-central-dp}, it remains to construct, for every $j$, a coupling between $P_{+j}$ and $P_{-j}$ whose expected Hamming distance is
small.

Fix $j$ and sample $V$ uniformly from $\{v:v_j=+1\}$ and define
$ V'=V^{(j)}.$ Then $V'$ is uniformly distributed over $\{v:v_j=-1\}$. Hence, if $Y\mid V=v \sim P_v, Y'\mid V'=v^{(j)} \sim P_{v^{(j)}},$ then the marginal distribution of $Y$ is $P_{+j}$ and the marginal distribution of $Y'$ is $P_{-j}$.

For each sample $i$, by Pinsker's inequality and the single-sample KL bound above,
$$
\|P_{\theta_v,i}-P_{\theta_{v^{(j)}},i}\|_{\rm TV} \le C|\alpha_i-\beta_i|  = C a |x_{ij}|.
$$
By the maximal coupling, we can have $ y_i\sim P_{\theta_v,i}, y_i'\sim P_{\theta_{v^{(j)}},i}$
such that $\mathbb P(y_i\neq y_i') = \|P_{\theta_v,i}-P_{\theta_{v^{(j)}},i}\|_{\rm TV} \le C a |x_{ij}|.$
Coupling coordinates independently over $i=1,\ldots,n$ gives
$$
D:=\mathbb E[d_{\rm Ham}(Y,Y')]= \sum_{i=1}^n \mathbb P(y_i\neq y_i') \le  C a\sum_{i=1}^n |x_{ij}| \le C n a\sqrt{\gamma},
$$
where the last inequality comes from Cauchy's inequality and Assumption~\ref{asm:boundness}.
Now apply the central-DP Assouad lemma~\ref{lem:Assouad-central-dp} with $d$ replaced by $m=s^*$,
$\alpha=2$, and $\tau=2a^2$. Then every central
$(\varepsilon,\delta)$-Label-DP estimator satisfies
$$
\inf_{\mathcal M\in\mathcal M_{(\varepsilon,\delta)}}
    \sup_{v\in\{-1,+1\}^m}
    \mathbb E_v\|\mathcal M(D)-\theta_v\|_2^2 \ge
    \frac{m\tau}{2\alpha}
    \left( 0.9 e^{-10\varepsilon D} - 10D\delta \right) =
    \frac{m a^2}{2}
    \left(0.9 e^{-10\varepsilon D} - 10D\delta\right).
$$
Choosing $a = \frac{c_1}{n\sqrt{\gamma}(\varepsilon+\delta)}$
for a sufficiently small universal constant $c_1>0$, we obtain
$$
\inf_{\mathcal M\in\mathcal M_{(\varepsilon,\delta)}}
    \sup_{v\in\{-1,+1\}^m}
    \mathbb E_v\|\mathcal M(D)-\theta_v\|_2^2
    \gtrsim
    m a^2 =
    \frac{m}{\gamma n^2(\varepsilon+\delta)^2} =
    \frac{s^*}{\gamma n^2(\varepsilon+\delta)^2}.
$$

\section*{Limitations}
This work focuses on sparse linear reward estimation under the BTL model, which provides a clean setting for theoretical analysis. Extending the framework to nonlinear or neural reward models, as well as to other forms of sparsity in representations or parameter-efficient modules, is an important future direction.

\section*{Broader Impact}
This work aims to improve privacy-preserving RLHF by reducing the privacy cost of reward estimation in high-dimensional settings. By exploiting sparsity, our methods may help make private alignment more practical when preference data are sensitive or expensive to collect, while still requiring standard safety and bias evaluations in real deployments.

\newpage
\input{checklist.tex}

\end{document}

%% file: checklist.tex
\section*{NeurIPS Paper Checklist}

\begin{enumerate}

\item {\bf Claims}
    \item[] Question: Do the main claims made in the abstract and introduction accurately reflect the paper's contributions and scope?
    \item[] Answer: \answerYes{} 
    \item[] Justification: Main contributions are stated in the abstract.

\item {\bf Limitations}
    \item[] Question: Does the paper discuss the limitations of the work performed by the authors?
    \item[] Answer: \answerYes{} 
    \item[] Justification: In remarks, we discussed the opening question that hasn't been addressed in this paper.

\item {\bf Theory assumptions and proofs}
    \item[] Question: For each theoretical result, does the paper provide the full set of assumptions and a complete (and correct) proof?
    \item[] Answer: \answerYes{} 
    \item[] Justification: All assumptions and proofs are provided.

    \item {\bf Experimental result reproducibility}
    \item[] Question: Does the paper fully disclose all the information needed to reproduce the main experimental results of the paper to the extent that it affects the main claims and/or conclusions of the paper (regardless of whether the code and data are provided or not)?
    \item[] Answer: \answerYes{} 
    \item[] Justification: All experimental information is provided,

\item {\bf Open access to data and code}
    \item[] Question: Does the paper provide open access to the data and code, with sufficient instructions to faithfully reproduce the main experimental results, as described in supplemental material?
    \item[] Answer: \answerYes{} 
    \item[] Justification: The full code will be released after the public version.

\item {\bf Experimental setting/details}
    \item[] Question: Does the paper specify all the training and test details (e.g., data splits, hyperparameters, how they were chosen, type of optimizer) necessary to understand the results?
    \item[] Answer: \answerYes{} 
    \item[] Justification: All details are stated.

\item {\bf Experiment statistical significance}
    \item[] Question: Does the paper report error bars suitably and correctly defined or other appropriate information about the statistical significance of the experiments?
    \item[] Answer: \answerYes{} 
    \item[] Justification: All results are discussed.

\item {\bf Experiments compute resources}
    \item[] Question: For each experiment, does the paper provide sufficient information on the computer resources (type of compute workers, memory, time of execution) needed to reproduce the experiments?
    \item[] Answer: \answerYes{} 
    \item[] Justification: All information is provided.
    
\item {\bf Code of ethics}
    \item[] Question: Does the research conducted in the paper conform, in every respect, with the NeurIPS Code of Ethics \url{https://neurips.cc/public/EthicsGuidelines}?
    \item[] Answer: \answerYes{} 
    \item[] Justification: Yes.

\item {\bf Broader impacts}
    \item[] Question: Does the paper discuss both potential positive societal impacts and negative societal impacts of the work performed?
    \item[] Answer: \answerYes{} 
    \item[] Justification: We provide a discussion in the appendix.
    
\item {\bf Safeguards}
    \item[] Question: Does the paper describe safeguards that have been put in place for responsible release of data or models that have a high risk for misuse (e.g., pre-trained language models, image generators, or scraped datasets)?
    \item[] Answer: \answerNA{} 
    \item[] Justification: The paper poses no such risks.

\item {\bf Licenses for existing assets}
    \item[] Question: Are the creators or original owners of assets (e.g., code, data, models), used in the paper, properly credited and are the license and terms of use explicitly mentioned and properly respected?
    \item[] Answer: \answerNA{} 
    \item[] Justification: All assets are cited.

\item {\bf New assets}
    \item[] Question: Are new assets introduced in the paper well documented and is the documentation provided alongside the assets?
    \item[] Answer: \answerNA{} 
    \item[] Justification: The paper does not release new assets.

\item {\bf Crowdsourcing and research with human subjects}
    \item[] Question: For crowdsourcing experiments and research with human subjects, does the paper include the full text of instructions given to participants and screenshots, if applicable, as well as details about compensation (if any)? 
    \item[] Answer: \answerNA{} 
    \item[] Justification: The paper does not involve crowdsourcing nor research with human subjects.

\item {\bf Institutional review board (IRB) approvals or equivalent for research with human subjects}
    \item[] Question: Does the paper describe potential risks incurred by study participants, whether such risks were disclosed to the subjects, and whether Institutional Review Board (IRB) approvals (or an equivalent approval/review based on the requirements of your country or institution) were obtained?
    \item[] Answer: \answerNA{} 
    \item[] Justification: The paper does not involve crowdsourcing nor research with human subjects.

\item {\bf Declaration of LLM usage}
    \item[] Question: Does the paper describe the usage of LLMs if it is an important, original, or non-standard component of the core methods in this research? Note that if the LLM is used only for writing, editing, or formatting purposes and does \emph{not} impact the core methodology, scientific rigor, or originality of the research, declaration is not required.
    \item[] Answer: \answerNA{}
    \item[] Justification: The core method development in this research does not involve LLMs as any important, original, or non-standard components.

\end{enumerate}